\documentclass[runningheads]{llncs}

\usepackage{eccv}

\usepackage{eccvabbrv}

\usepackage{graphicx}
\usepackage{booktabs}

\usepackage[accsupp]{axessibility}  % Improves PDF readability for those with disabilities.

\usepackage{hyperref}

\usepackage{orcidlink}
\usepackage{array}
\usepackage{multicol}
\usepackage{multirow}
\newcommand{\methodname}{\textbf{AnyHome}}

\usepackage{algorithmic}
\usepackage{algorithm}
\usepackage{tabularx}
\usepackage{mathrsfs}

\begin{document}

% ---------------------------------------------------------------
% TODO REVIEW: Replace with your title
\title{\methodname{}: Open-Vocabulary Generation of Structured and Textured 3D Homes} 

% TODO REVIEW: If the paper title is too long for the running head, you can set
% an abbreviated paper title here. If not, comment out.
\titlerunning{\methodname{}}

% TODO FINAL: Replace with your author list. 
% Include the authors' OCRID for the camera-ready version, if at all possible.
\author{Rao Fu\textsuperscript{*†}\inst{1}\orcidlink{0000-0002-0115-0831} \and
Zehao Wen\textsuperscript{*}\inst{2}\orcidlink{0009-0003-7385-3397} \and
Zichen Liu\textsuperscript{*}\inst{2}\orcidlink{0009-0005-6344-2425} \and
Srinath Sridhar\inst{1}\orcidlink{0000-0003-4663-3324}
}

% TODO FINAL: Replace with an abbreviated list of authors.
\authorrunning{R. Fu et al.}
% First names are abbreviated in the running head.
% If there are more than two authors, 'et al.' is used.

% TODO FINAL: Replace with your institution list.
\institute{Brown University \and
Shenzhen College of International Education \\
\textsuperscript{*}Equal Contribution. \textsuperscript{†} Corresponding Author.\\
\url{https://ivl.cs.brown.edu/research/anyhome} 
}

\maketitle

\begin{figure}[t]
\begin{center}
\includegraphics[width=0.98\linewidth]{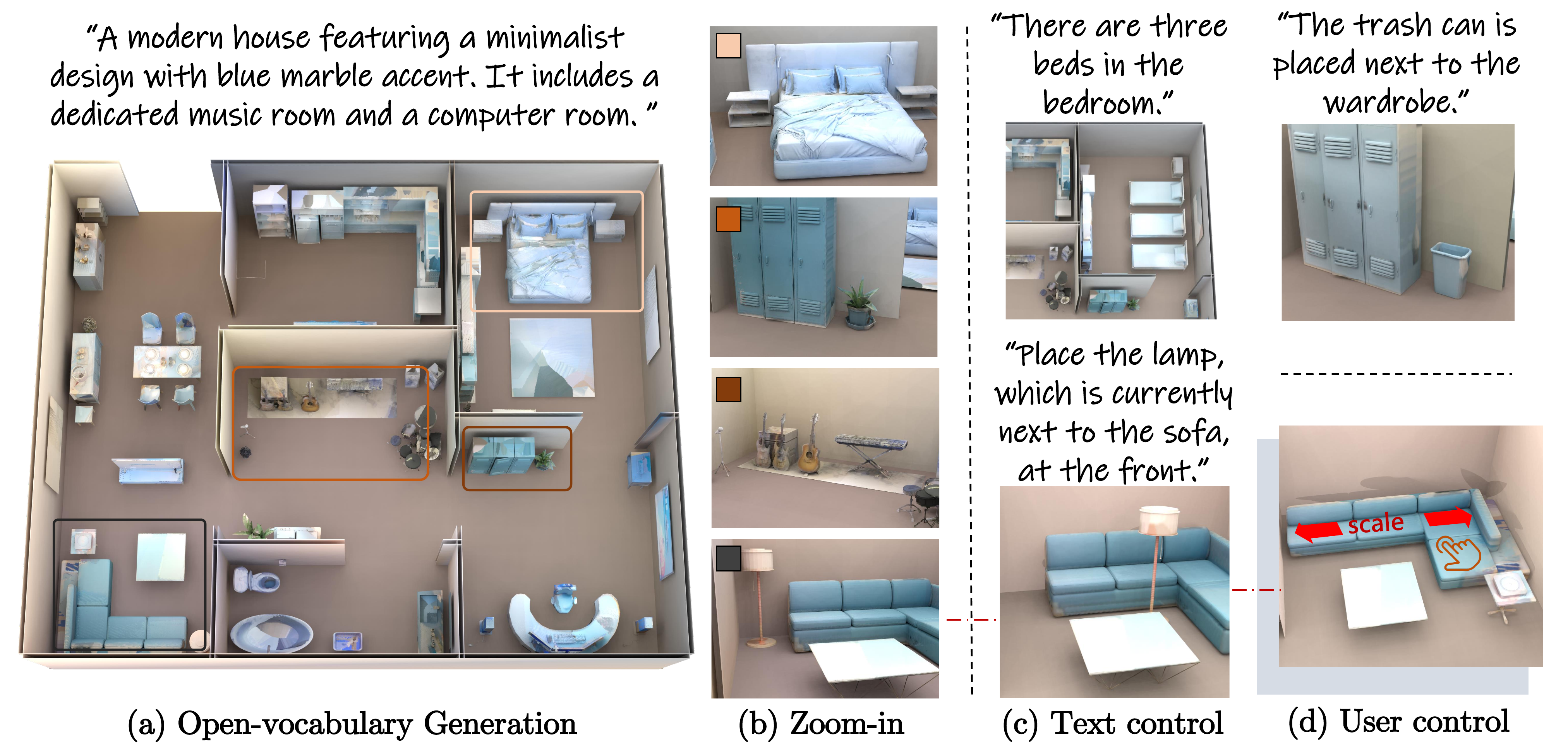}
\end{center}
\caption{\textbf{Example house-scale indoor scene generated by \methodname{}.} Users can input any textual description of an indoor scene, and the system is capable of generating house floorplans, room layouts, object placements, and stylistic appearances accordingly. The generated indoor scene is represented by structured and textured room and object meshes. \methodname{} enables the synthesis of diverse indoor scenes, allowing users to control scene generation at any stage—from textual input and intermediate representation to the generated meshes. }
\label{fig:teaser}
\end{figure}

\begin{abstract}
Inspired by cognitive theories, we introduce \methodname{}, a framework that translates any text into well-structured and textured indoor scenes at a house-scale. By prompting Large Language Models (LLMs) with designed templates, our approach converts provided textual narratives into amodal structured representations. These representations guarantee consistent and realistic spatial layouts by directing the synthesis of a geometry mesh within defined constraints. A Score Distillation Sampling process is then employed to refine the geometry, followed by an egocentric inpainting process that adds lifelike textures to it. 
\methodname{} stands out with its editability, customizability, diversity, and realism. The structured representations for scenes allow for extensive editing at varying levels of granularity. Capable of interpreting texts ranging from simple labels to detailed narratives, \methodname{} generates detailed geometries and textures that outperform existing methods in both quantitative and qualitative measures.
\keywords{Scene Generation \and 3D Synthesis \and Vision and Language}
\end{abstract}
\section{Introduction}
Homes are pivotal to our existence. They are the silent witnesses to our routines and landmark moments, repositories of personal and collective memories that influence our well-being and behavior. Imagine the possibilities if we could articulate our ideal living spaces in natural language and see them come to life. \methodname{} embodies this vision, offering a framework that transforms free-form, open-vocabulary textual narratives into diverse, house-scale 3D indoor scenes with realistic appearances. These scenes can be used in a variety of domains, including interior design, game development, augmented and virtual reality, as well as the training for embodied agents. They feature intricate structure layouts and naturalistic textures that are readily modifiable, bridging a crucial gap in contemporary digital design practices.

Previous research in text-guided 3D scene generation \cite{fridman2023scenescape, hollein2023text2room, huang2023aladdin, tang2023mvdiffusion} has made notable advances but often struggles with creating robust 3D structures, sometimes resulting in rooms with open ends or repetitive layouts. Studies focusing on structured scene generation \cite{HouseGAN, HouseGAN++, ATISS, LegoNet, tang2023diffuscene} are typically confined to predefined room and furniture types, lacking the flexibility to accommodate customization for various room types, furniture pieces, small objects, and their respective arrangements. At the same time, the generation of house-scale scenes with diverse structures is crucial, especially for applications requiring seamless navigation across different rooms, like video games or virtual training for embodied agents. Some studies have explored house-scale scene synthesis \cite{deitke2022️}, but they often rely on predefined artificial textures, which may limit controllability and realism, hindering certain downstream tasks \cite{khanna2023habitat}.

In response to these limitations, \methodname{} focuses on creating indoor scenes that are customizable through \textbf{open-vocabulary} text inputs, featuring \textbf{structured representations}, \textbf{realistic texture}, and are \textbf{scalable to house-size}.
We draw inspiration from two key hypotheses in environmental cognition. The \textit{amodal spatial image} hypothesis suggests that environment recognition transcends sensory modalities like vision or hearing, proposing that individuals apprehend their surroundings through an abstract, symbolic, map-like representation \cite{loomis2002spatial, giudice201815}. Conversely, the \textit{visual recording} hypothesis posits that visual experiences function similarly to a video camera, perceiving the environment through continuous, egocentric, path-oriented exploration \cite{pick1974visual, gibson2014ecological}. Learning from both concepts, \methodname{} maintains an amodal, hierarchical geometry representation during the generation process to conceptualize the environment's structure. To enhance visual realism, we adopt an egocentric exploration approach for in-painting, encouraging the model to detail the environment as it ``sees''.

Specifically, the scene generation process in \methodname{} unfolds in several stages: textual input modulation, hierarchical structured geometry generation, along with egocentric refinement and inpainting. This approach is markedly text-controllable. Users can design a scene through free-form text descriptions, which are subsequently converted into modular descriptions for aspects such as floorplans, room layouts, object retrieval and placement, as well as scene appearances, all facilitated by querying Large Langueg Models (LLMs) with specialized templates. These descriptions transformed to graph-based intermediate representations, which fosters customizability and solves the problem of insensible geometry generated directly from the LLMs. 
Following this, the scene's floorplan is obtained through a generative model \cite{HouseGAN, HouseGAN++}, and room layouts together with object placements are determined according to LLM-dictated placement rules. We use databases for furniture\cite{3DFuture} and objects\cite{deitke2023objaverse} to populate these scenes.
During the egocentric inpainting phase, a camera trajectory imitating first-person exploration is generated. Textures are painted along this trajectory using a depth-aware, text-conditioned inpainting model, which aligns the texture with existing geometry \cite{tang2023mvdiffusion}.
However, we note inaccuracies in furniture and object placements due to the inherent limitations in object canonicalization from the mesh datasets. To counter, we integrate a Score Distillation Sampling (SDS) process with a differentiable renderer to refine object placement, inspired by recent 3D generation advancements using SDS loss\cite{poole2022dreamfusion} for geometry optimization.

Our experiments demonstrate that \methodname{} effectively generates house-scale scenes with compelling structures, visually appealing textures, and strong alignment with provided textual inputs. The language accessibility and structured representations enable users to control scene generation at various levels, from room types and layouts to object placements and appearances. 

To summarize, our key contributions include:
\begin{enumerate}

\item Developing a systematic and reliable approach for creating diverse, text-controlled, texture-realistic, and modifiable scenes at a house-scale, catering to a wide range of applications.
\item Utilizing LLMs to convert open-vocabulary textual input into structured representations, allowing detailed control using language-based customization while maintaining structure consistency.
\item Enhancing object placement using an SDS process, thus increasing the system's robustness and versatility in creating sensible scenes.
\item Innovating an egocentric inpainting process that follows a camera trajectory generated automatically to explore each object in the room.

\end{enumerate}

\section{Related Work}
\noindent \textbf{House-scale Floorplan Generation.}
House floorplan generation has been well-studied through methods like shape grammars and iterative generation processes\cite{ma2014game, merrell2010computer, muller2006procedural, peng2014computing}. With the advent of deep learning, the focus has shifted towards using graph-constrained generative networks\cite{hu2020graph2plan, sun2022wallplan, luo2022floorplangan, bisht2022transforming, HouseGAN, HouseGAN++, shabani2023housediffusion, tang2023graph}, accommodating a variety of user-defined constraints. As our work concentrates on the functionality of rooms, we utilize bubble diagrams as a constraint\cite{HouseGAN++}, which outline pre-defined room types and the functional connections between them. Our primary contribution lies in synthesizing bubble diagrams based on user textual input and expanding these diagrams to encompass any room type, beyond the predefined categories.

\noindent \textbf{Room-scale Scene Generation.}
Room-scale scene generation involves creating 3D content, such as furniture and smaller objects, and determining their placement. Traditional methods typically start by constructing a set of 3D objects and then optimizing their placement using iterative methods\cite{fisher2015activity, fu2017adaptive, deitke2022️}, non-linear optimization\cite{qi2018human, yu2011make, xu2013sketch2scene, fisher2012example}, or manual interaction\cite{merrell2011interactive, huang2023aladdin}. However, with the development of large-scale indoor scene datasets\cite{paschalidou2021atiss, 3DFront, yeshwanthliu2023scannetpp}, recent approaches have shifted towards using generative networks, employing techniques like feed-forward networks\cite{zhang2020deep}, VAEs\cite{purkait2020sg}, GANs\cite{yang2021indoor}, autoregressive models\cite{wang2018deep, wang2019planit, wang2021sceneformer, li2019grains}, and diffusion models\cite{LegoNet, tang2023diffuscene, zhai2023commonscenes}. These advanced methods can generate diverse and realistic 3D scenes but often struggle to place objects unpresented in the training dataset. Recent studies\cite{LayoutGPT, yang2024holodeck} have attempted to overcome these limitations by using Large Language Models (LLMs) to generate style sheet languages for object placement. Our method generates room layouts in two steps. We initially create 3D layouts using LLMs in a few-shot manner and then employ a differentiable renderer to refine the placement of objects.

\noindent \textbf{Text-to-Shape Generation.}
Text-to-Shape Generation has seen considerable development, with many studies utilizing feed-forward methods\cite{achlioptas2018learning, chen2019text2shape, mittal2022autosdf, liu2022towards, sanghi2022clip, fu2022shapecrafter, cheng2023sdfusion,sanghi2023clip, wei2023taps3d, nichol2022point, jun2023shap} to train generators on 3D data. These methods generate 3D shapes efficiently but their ability to generalize is often limited by the size of the 3D data on which they are trained. Some recent approaches\cite{jain2022zero, mohammad2022clip, poole2022dreamfusion, lin2023magic3d, wang2023prolificdreamer} leverage pre-trained visual-language knowledge to optimize 3D representations, showing promising generalization to open-vocabulary textual inputs. Rather than generating shapes from scratch, some research\cite{richardson2023texture, ma2023x, chen2023text2tex, cao2023texfusion, hwang2023text2scene} focuses on texture generation, aiming to create textures represented as UV maps from a given mesh. However, optimizing camera views for a single shape is relatively straightforward and cannot directly translate to the complexity of scene geometry. Our paper introduces a novel method for generating camera trajectories specifically for texturing more complex and varied scene structures.

\noindent \textbf{Text-to-Scene Generation.}
Text-to-Scene Generation has followed two primary approaches. The first approach\cite{chang2017sceneseer, wang2021sceneformer, tang2023diffuscene} directly uses 3D representations for generation. These methods begin by transforming user text inputs into graph-based representations or lists of shape codes, subsequently generating 3D scenes based on these intermediate representations. While effective in creating structured 3D representations, they often fail to generate realistic textures and unseen objects. The second approach addresses these issues through the image domain. Some studies\cite{chen2022text2light} use panorama images for scene representation and develop end-to-end models that convert text to panorama for scene generation. Other research \cite{bautista2022gaudi, fridman2023scenescape, hollein2023text2room, song2023roomdreamer, bahmani2023cc3d} explores scenes from first-person perspectives, employing large-scale text-conditioned image generation models for appearance creation. Techniques like neural density fields or depth are used to ensure 3D consistency. This approach excels in generating diverse appearances but struggles with creating structured 3D representations and consistent shapes. Our method overcomes these challenges by adopting a two-stage generation process. We first generate spatial layouts in a few-shot manner and then inpaint textures using first-person viewpoints, which guarantees both structured outputs and diverse appearances.
\begin{figure*}[t]
\begin{center}
\includegraphics[width=1.0\textwidth]{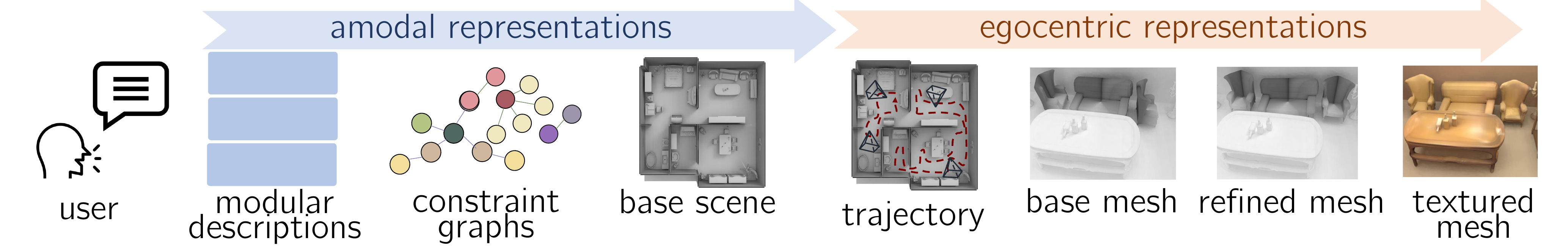}
\end{center}
\caption{\textbf{Two-stage Generation Process.} \methodname{} unfolds two primary steps: First, amodal representations are generated from user's text input, which involves constructing modular text descriptions, constrained graphs, and hierarchical structured base mesh. Following this, the method embarks on an egocentric exploration stage, where a navigation trajectory is generated, enabling the refinement and texturing of the base mesh from different viewpoints.}
\label{fig:overall}
\end{figure*}
% \vspace{-0.5cm}

\section{\methodname{}}
\label{sec:method}
\begin{figure*}[t]
\begin{center}
\includegraphics[width=1.0\textwidth]{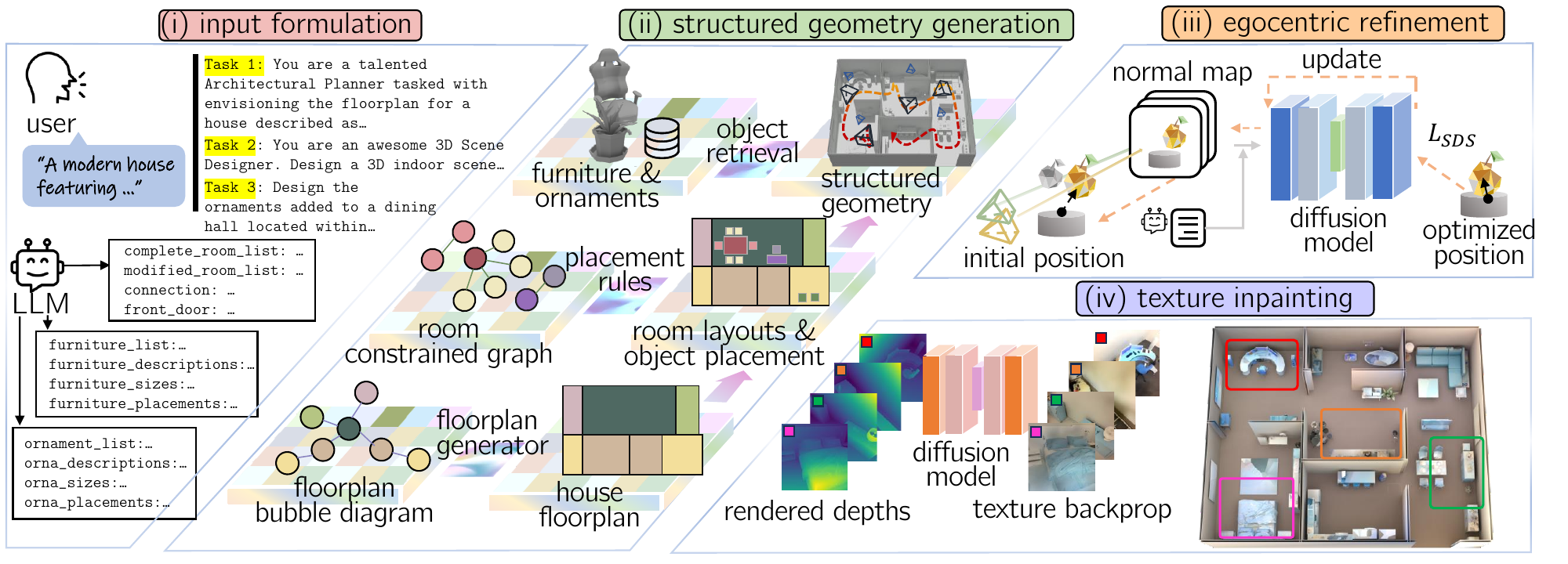}
\end{center}
\caption{\textbf{Pipeline.} Taking a free-form textual input, our pipeline generates the house-scale scene by: (i) comprehending and elaborating on the user's textual input through querying an LLM with templated prompts; (ii) converting textual descriptions into base geometry using structured intermediate representations; (iii) employing an SDS process with a differentiable renderer to refine object placements; and (iv) applying depth-conditioned texture inpainting for egocentric texture generation.}
\label{fig:pipeline}
\end{figure*}

\cref{fig:overall} illustrates the two-stage generation process of \methodname{}. Our approach incorporates insights from environmental cognition hypothesis. We adopt the \textit{amodal spatial image} hypothesis\cite{loomis2002spatial,giudice201815} to represent houses with an amodal hierachical structured representation. Drawing on the \textit{visual recording} hypothesis\cite{pick1974visual,gibson2014ecological}, we further refine and enrich the visual details of houses using egocentric inpainting. \cref{fig:pipeline} illustrates the main components of the pipeline. Our framework comprises three main components: First, we modulate the textual input using Large Language Models (LLMs) as detailed in \cref{sec:text}, which involves comprehending and elaborating user input. Second, in \cref{sec:structure}, we focus on synthesizing graph-based structured representations that are important for controllability, and use them to generate the base geometry. Finally, \cref{sec:egocentric} describes our process of refinement and inpainting through egocentric exploration, which refines object placements and adds texture, enhancing the realism of the generated scene.
% \vspace{-0.5cm}
\subsection{Textual Input Modulation.}
\label{sec:text}
This initial component of our framework performs two key functions. First, it allows users to offer an unrestricted description of the scene to guide and control the generation of designs. Second, it utilizes the common-sense knowledge inherent within LLMs \cite{LLMSurvey, LLMEmergent} to convert the provided input into modular descriptions for house floorplans, room layouts, and appearances of furnishings and ornaments. Notably, LLMs are instructed to render graph-like structures that act as intermediate constraints for the generation of house floorplans and room layouts. We have designed three sets of prompts to guide the LLMs: for house floorplan generation (\(p_{\text{floorplan}}, p_{\text{map}}\)), for room layout and object placement (\(p_{\text{room}}\)), and for room appearance (\(p_{\text{appearance}}\)). \cref{fig:pipeline} (i) provides an example of using (\(p_{\text{room}}\)) to generate room layout descriptions. The design of these prompts bases on three principles: format compatibility with subsequent modules, adherence to user's textual specifications, and maximal detail elaboration by the LLM-agent, which is encouraged to adopt the perspective of an interior designer or navigator. Additionally, numerical data provided to LLMs is converted to a real-world scale (in meters) to fully utilize their knowledge base. Further details on these prompts can be found in the Supplementary Material.

\subsection{Hierarchical Structured Geometry Generation.}
\label{sec:structure}
In this phase, we concentrate on converting modular descriptions into a detailed base geometry of the scene, covering elements like floors, walls, furniture, and smaller objects. For precise spatial control, we employ two graph-based intermediate representations: one dedicated to the floorplan and the other to room layouts. \cref{fig:pipeline} (ii) illustrates this hierarchical generation process.

\subsubsection{House Floorplan Generation.}

A floorplan is crucial as it outlines the spatial arrangement, dimensions, and functionalities of rooms. While zero-shot methods like LayoutGPT\cite{LayoutGPT} can generate a variety of room types, they sometimes yield impractical configurations, as illustrated in \cref{fig:result_comparison}. To address these challenges, we initially create bubble diagrams, a graph-based representation similar to that in \textit{HouseGAN++}\cite{HouseGAN++}, and input them into pre-trained floorplan generation networks to facilitate text-controlled floorplan synthesis.

In this approach, each floorplan is represented by a bubble-diagram $G(N, E)$, where nodes $N$ symbolize room types and edges $E$ represent functional connections. We query an LLM with a specific floorplan generation prompt $p_{\text{floorplan}}$ to convert user inputs into these diagrams. Utilizing the pre-trained, graph-conditioned floorplan generator \textit{HouseGAN++}, we transform these diagrams into a set of masks $M$ that detail the spatial layout.

Considering \textit{HouseGAN++}'s limitations with its \textit{RPLAN} training data, we introduce additional prompts $p_{\text{map}}$ to map unconventional room types to \textit{RPLAN}'s recognized categories. Upon defining $M$, we construct the house's base mesh at a defined height $h$. This method, combining LLMs with a specialized floorplan network, maintains variety while ensuring spatial realism.

\subsubsection{Room Layout and Object Placement Generation.}

With the house floorplan established, we advance to generating room layouts and object placements. While existing methods have shown limitations with unique room types\cite{ATISS, LegoNet} and simplistic layouts\cite{LayoutGPT} as evidenced in \cref{fig:result_comparison}, our approach introduces a constrained layout graph $\bar{G_i}(\bar{N_i},\bar{E_i})$ to enhance coherence and flexibility. This graph captures dimensions and descriptions of objects in its nodes $\bar{N_i}$ and the relationships between objects in edges $\bar{E_i}$.

Using LLMs, we generate a graph $\bar{G_i}$ for each room based on its area and type, as well as the user input. LLMs struggle to envision a room with a diverse array of objects and to manage their complex interrelations simultaneously, and hence we prompt the LLM twice during this graph generation process. We employ two distinct prompt templates: $p_\text{furniture}$ for large furniture and $p_\text{ornament}$ for smaller ornaments, to address these elements separately.
 
The features stored in the nodes and edges are generated in a step-by-step manner. For instance, in furniture placement, we initially generate a list of potential furniture items along with their dimensions. Subsequently, drawing on the concept of Semantic Asset Group (SAG)\cite{deitke2022️}, the LLM is prompted to identify the sub-graphs $C_i$ within $\bar{G_i}$ that represent groups of commonly associated items and designate an anchor piece within each group, such as a table in a set of a table and chairs.

With a predefined set of placement rules like \textit{place\_corner($\cdot$)}, \textit{place\_beside($\cdot$)}, and \textit{place\_wall($\cdot$)}, we direct the LLM to determine the placement rules between the anchor object and other items within the SAG. The corresponding placement algorithm for each rule then dictates the objects' locations, orientations, and adjustments in the event of obstructions, thus optimizing space usage. The Supplementary Material provides a detailed account of all placement rules, including their functions and parameters.
This stage results in a series of bounding boxes $B_i$ outlining the final layout.

\subsubsection{Object Retrieval.}
To enhance the scene with appropriate furniture and decor, we utilize comprehensive mesh datasets\cite{3DFuture, deitke2023objaverse}, retrieving items that best match the LLM-generated descriptions. By employing CLIP\cite{CLIP} and Sentence Transformers\cite{reimers-2019-sentence-bert}, we ensure that the retrieved items align closely with the textual descriptions, seamlessly integrating them into the overall house mesh.

\subsection{Egocentric Refinement and Inpainting.}
\label{sec:egocentric}
In this stage, we enhance the base geometry in an egocentric manner, refining its details and layouts as well as adding textures. This process is illustrated in \cref{fig:pipeline} (iii) and (iv). We start by creating an egocentric trajectory, followed by layout refinement using Score Distillation Sampling (SDS) loss. The final step involves using a depth-conditioned inpainting model for texturing, ensuring coherence between the geometry and textures. Also, a differentiable renderer is employed to maintain texture consistency from multiple viewpoints.

\subsubsection{Generating Egocentric Trajectories.}
Our trajectory generation adheres to three principles: (1) ensuring complete coverage by guiding the camera along walls; (2) focusing on object-centric views, facilitated by our structured geometry; and (3) implementing double traversal to capture views both towards and away from the interior walls. Additionally, we include random camera samples to achieve comprehensive coverage. This process begins by creating an obstacle field based on the base geometry. By setting a specific threshold, we can identify key points within the field at certain magnitudes to define the trajectory, with the gradient indicating the camera's direction. For random sampling, we select from a set of closed-loop trajectories as feature points and employ a Bezier curve to smoothly interpolate the camera path. Detailed information on the algorithm and its visualizations can be found in the Supplementary Material.

\subsubsection{Refinement and Inpainting with Text-to-Image Models.}
With the egocentric camera trajectories set, we still face challenges: un-textured geometry and layout inaccuracies resulting from placement rules and inconsistent object coordinate systems from the mesh datasets. We address these by optimizing object placement and texture with image domain losses. In a given scene with $N$ meshes, denoted as $M=\{m_i | i \in [1, N]\}$, the position of each mesh $P=\{p_i | i \in [1, N]\}$ is defined relative to its anchor mesh, parameterized as the relative translation and quaternion rotation. The corresponding vertex colors for these meshes are symbolized by $M^{c}=\{m^{c}_i | i \in [1, N]\}$. Following Fantasia3D \cite{chen2023fantasia3d}, we divide the process into two distinct stages: placement refinement and mesh texturing.

\textbf{Structure Refinement.} We render the mesh set's normal map $(n,o)$ and mask $(o)$ from camera views $(c)$ using a differentiable renderer $(\Psi)$. The position parameters $P$ are then refined using the SDS loss, calculated as follows:
\begin{equation}
    \nabla_{P} \mathcal{L}_\text{SDS}(\phi, \Psi(M,c)) = 
    \mathbb{E}_{t, \epsilon} \!\! \left[ w(t)(\hat{\epsilon}_{\phi}(\tilde{n}_t;y,t) \!-\! \epsilon)\frac{\partial z}{\partial P}\! \right],
\end{equation} 
where $\phi$ parameterizes the pre-trained stable diffusion model, and $\hat{\epsilon}_{\phi}(z_t^{\tilde{n}}; y, t)$ represents the noisy function given the noisy image $\tilde{n}_t$, text embedding $y$, and noise level $t$, with $\epsilon$ being the added noise. We utilize mutli-view images with a batch size of $8$ for optimization. This refinement improves object positioning, surface adherence, as well as penetration issues.

\textbf{Texturing with Depth-conditioned Inpainting.}
To ensure realistic generation and preservation of 3D geometry, depth maps $(\text{d}_i)$ are rendered from each viewpoint. The LLM transforms user textual input $(t)$ into an appearance diffusion prompt $(p_{\text{diffusion}})$ to the diffusion model, utilizing a predefined appearance prompt $(p_{\text{appearance}})$. This diffusion prompt directs a depth-to-image inpainting model to produce textured images $(I_i = \phi(p_{\text{diffusion}}, d_i))$. These images are then back-projected into 3D space, iteratively refining the mesh texture as $M^{c_{i+1}} = \Psi(M^{c_i}, I_i)$. This approach ensures the alignment of the textured mesh with the original 3D geometry and its detailed design elements, such as adding calligraphy to scrolls.

\section{Results}
\label{sec:results}
In this section, we present our results on open-vocabulary 3D house generation and editing. We also provide comparison with other methods both quantitatively and qualitatively.

\begin{figure}[t]
\begin{center}
\includegraphics[width=\textwidth]{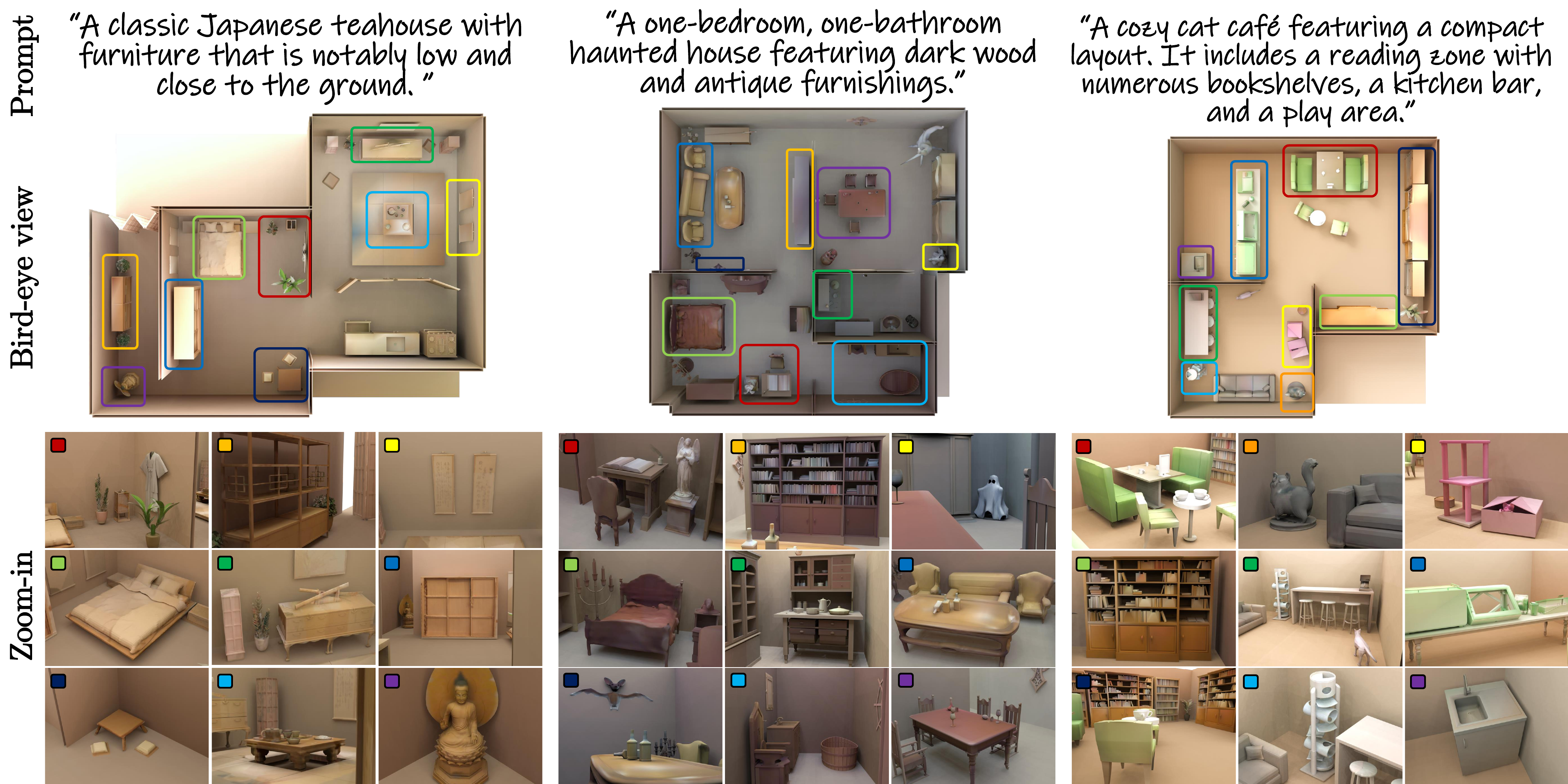}
\end{center}
% \vspace{-0.5cm}
\caption{\textbf{Open-Vocabulary Generation Results.} Top: Input text prompt. Middle: Bird's-eye view of the scenes. Bottom: Egocentric view of the scenes. \methodname{} interprets users' textual inputs and produces structured scenes with realistic textures. It can create a serene and culturally rich environment (Left - "Japanese tea house"), render a more dramatic and stylized ambiance (Middle - "haunted house"), and synthesize unique house types (Right - "cat cafe").}
\label{fig:result_3D}
\end{figure}

\subsection{Open-vocabulary Scene Generation.}
\cref{fig:result_3D} illustrates \methodname{}'s ability to generate house-scale 3D scenes from open-vocabulary, free-form text. The framework interprets and elaborates on a wide array of styles, including both eastern and western designs. This allows for the creation of stylistically coherent rooms with appropriately placed objects. One notable aspect of \methodname{} is its capacity to generate an extended range of room types, such as tea rooms, moving beyond the limitations of the standard RPLAN list\cite{RPLAN}. 
The showcased examples are a testament to \methodname{}'s proficiency in creating various stylistic scenes, each distinguished by their coherent floorplans, layouts, and the high-quality, consistent textures that resonate with the specified textual descriptions.

\begin{figure}[t]
\begin{center}
\includegraphics[width=0.98\textwidth]{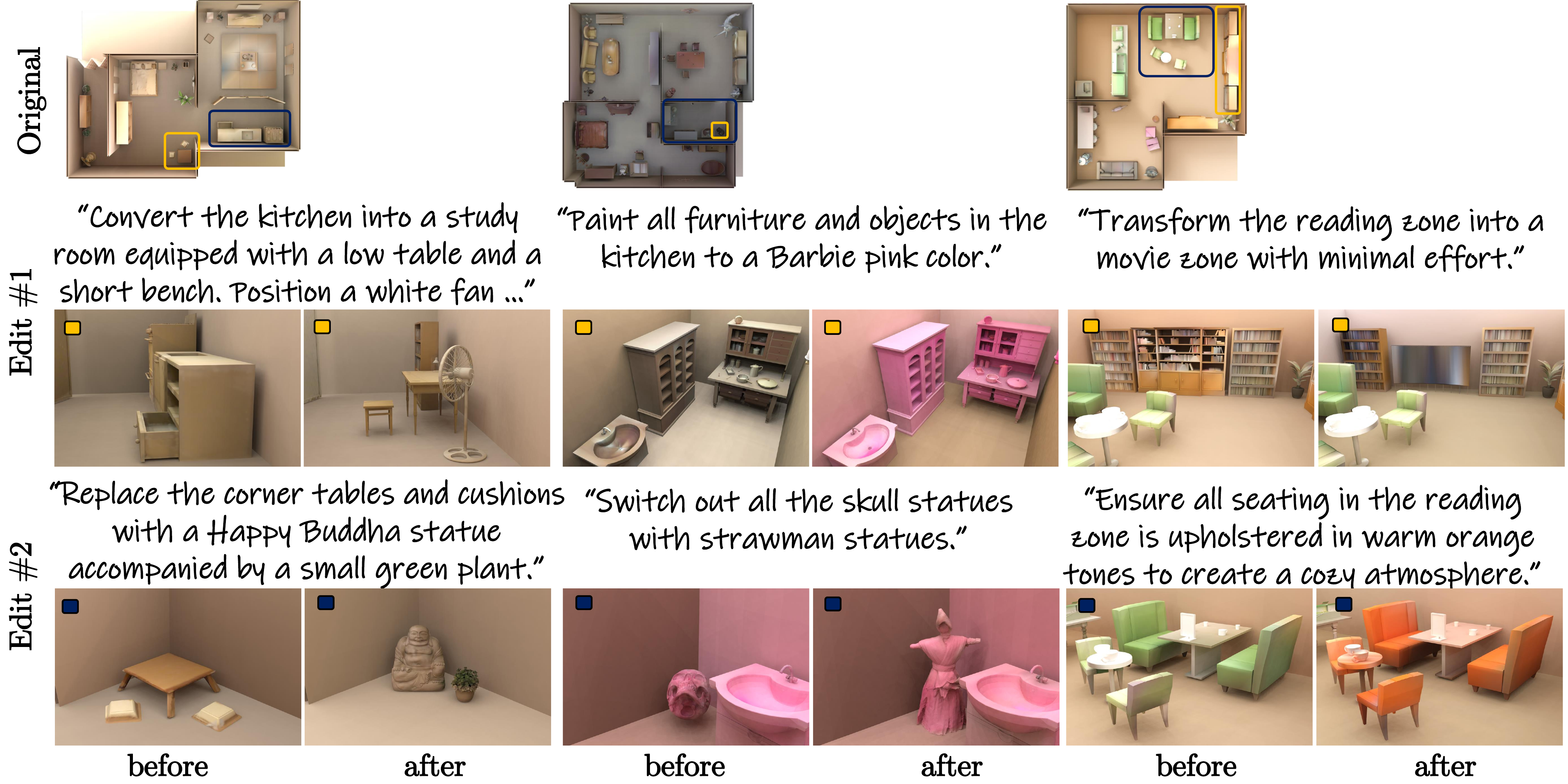}
\end{center}
% \vspace{-0.5cm}
\caption{\textbf{Open-vocabulary Editing Results.} Examples showcase \methodname{}'s capability to modify room types, layouts, object appearances, and overall design through free-form user input. \methodname{} also supports comprehensive style alterations and sequential edits, all made possible by its hierarchical structured geometric representation and robust text controllability.}
\label{fig:edit_3D}
\end{figure}

\subsection{Open-vocabulary Editing.}
We present a series of detailed editing results to further illustrate the customizability of \methodname{} in \cref{fig:edit_3D}. These examples highlight the framework's text-controllability across various levels: from altering room types (left row 1 - from kitchen to study room), adjusting room layouts (left row 2 -  from cushion set to a Buddha statue), modifying room styles (middle row 1 - from haunted to Barbie pink), to changing object appearances (right row 2 - from green seats to orange seats). Additionally, \methodname{} supports free-form design modifications (right row 1 -from reading to movie area) and facilitates sequential editing (middle row 2). These functionalities are made possible by the system's modular text prompt design and its hierarchical, structured geometry representations.

\begin{figure}[h]
\begin{center}
\includegraphics[width=\textwidth]{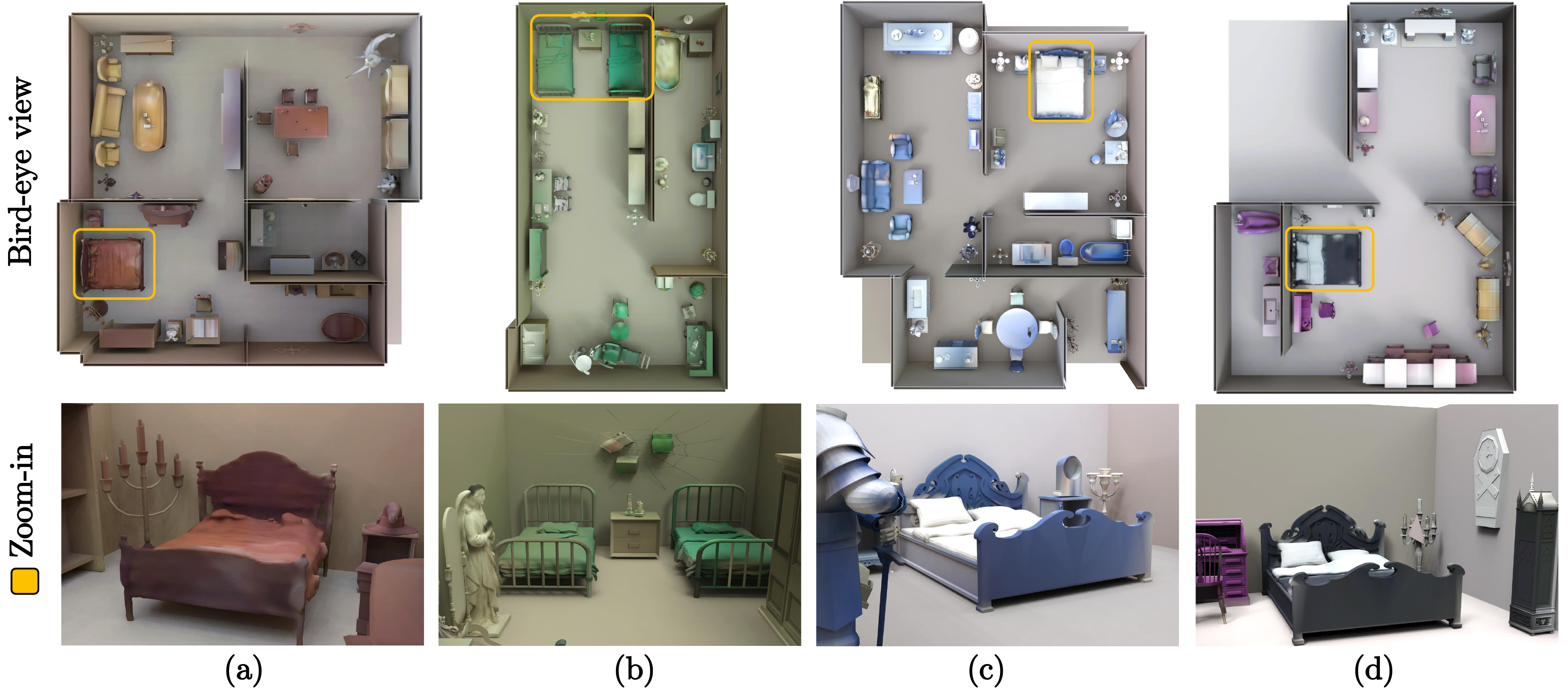}
\end{center}
\vspace{-0.5cm}
\caption{\textbf{Diverse Scene Results.} Four distinct scenes generated for the prompt "\texttt{A one-bedroom, one-bathroom haunted house featuring dark wood and antique furnishings.}" \methodname{} produces houses with diverse floorplans, room types, room layouts, objects and textures. }
\label{fig:result_diverse}
\end{figure}
\begin{figure}[t]
\begin{center}
\includegraphics[width=\textwidth]{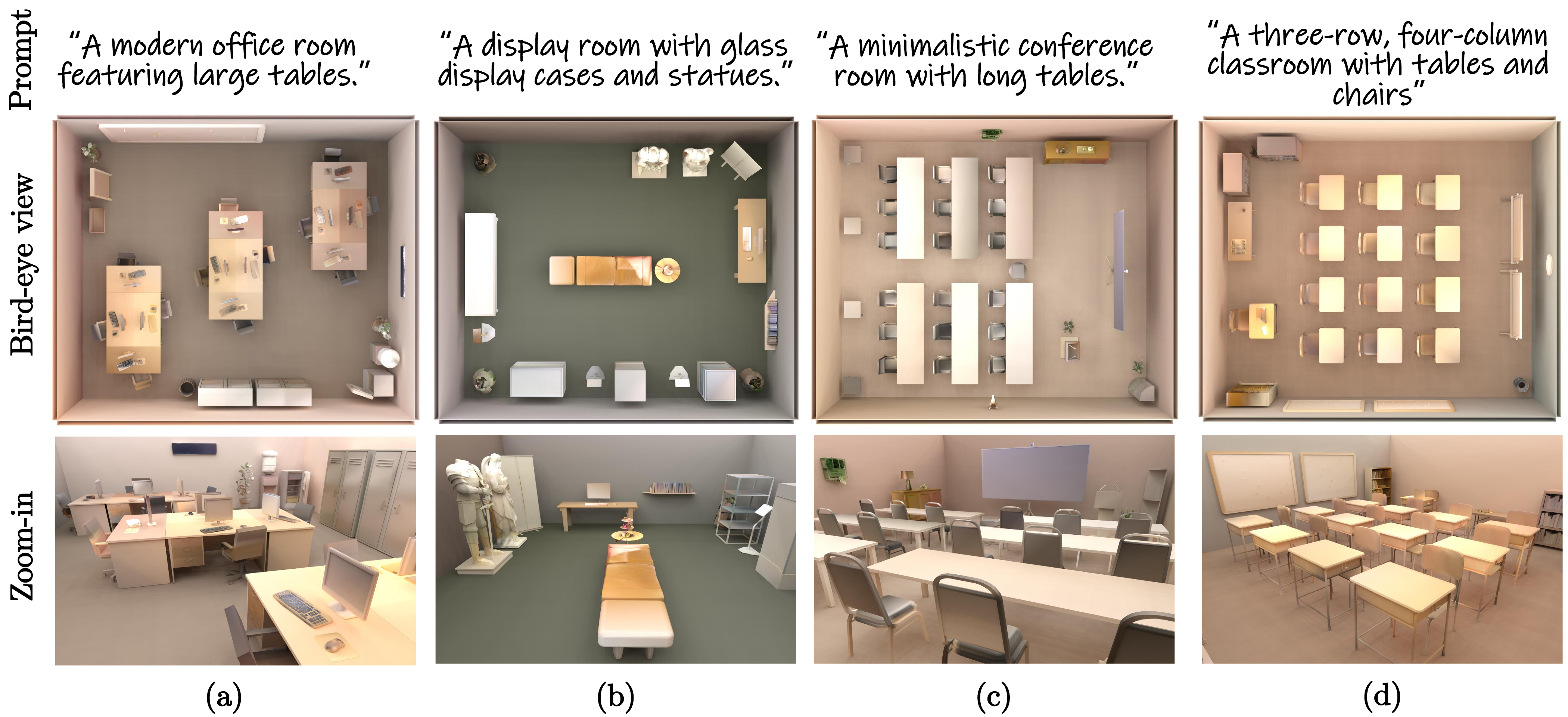}
\end{center}
\caption{\textbf{Diverse Layout Results.} \methodname{} can generate diverse layouts beyond usual room types, showcasing its versatility to be used for different applications. The layouts synthesized can be neatly symmetrical and orderly, echoing to the specifics provided in the prompt.}
\label{fig:result_layout}
\end{figure}

\subsection{Diversity.}
\cref{fig:result_diverse} showcases four unique scenes generated by \methodname{} from a single text input, demonstrating its capability for diversity. \methodname{} successfully generates varied floorplans (a five-room scene in \textbf{a} and a three-room scene in \textbf{c}), room types (a laboratory in \textbf{b}), room layouts (two parallel beds in \textbf{b}), room objects (a soldier in \textbf{c}), and appearances (different styles and colors), all while maintaining coherence with the same textual description. \cref{fig:result_layout} highlights \methodname{}'s ability to create layouts beyond traditional "home-like" environments. The system can produce various room types such as offices, display rooms, conference rooms, and classrooms with neatly symmetrical patterns.

\begin{figure}[t]
\begin{center}
\includegraphics[width=0.98\textwidth]{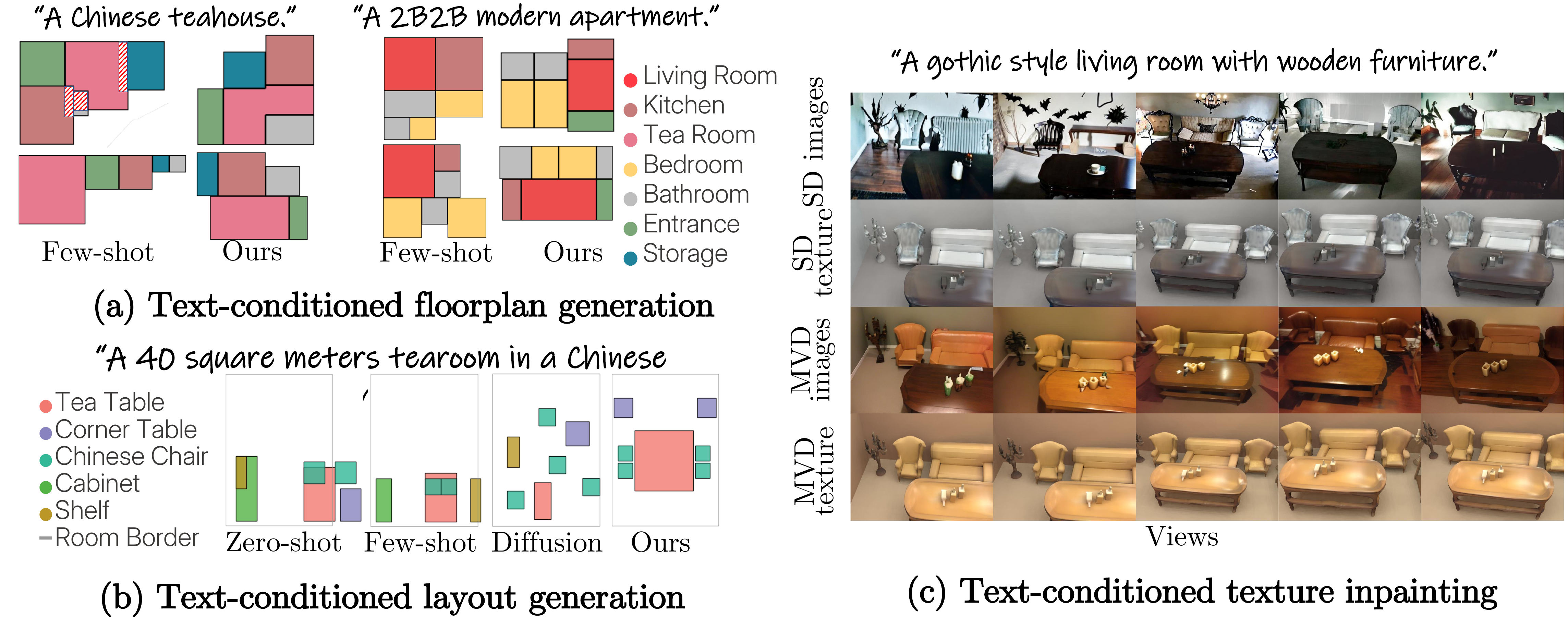}
\end{center}
\vspace{-0.5cm}
\caption{\textbf{Comparison of Structure Generation and Texture Inpainting.} For structure generation, our method generates more complex and reasonable floorplans and layouts. For texture inpainting, our method generates view-consistent and text-content aligned texture.}
\label{fig:result_comparison}
\end{figure}

\subsection{Comparison.}

We feature quantitative and qualitative comparisons of \methodname{} against other baseline methods in the following section. We provide comparisons with concurrent works in the Supplementary Material.

\subsubsection{Quantitative Comparison with Baselines}
To evaluate the effectiveness of \methodname{} in scene layout generation and text-to-scene alignment, we conduct a comparative analysis using several key metrics. \textbf{(1)} \textit{Out of Bound Rates (OOB):} The quality of layouts is quantified by measuring the frequency of objects intersecting with each other and the walls, or extending beyond room boundaries, following the approach in LayoutGPT\cite{LayoutGPT}.
\textbf{(2)} \textit{Caption Similarity (Caption-\textit{sim}):} This metric assesses the alignment between text and scene content. It involves generating captions\cite{li2023blip} for rendered frames and computing sentence similarity\cite{reimers-2019-sentence-bert} with the diffusion model $p_\text{diffusion}$. This is crucial for evaluating how well the layout and appearance match the user's input from an egocentric viewpoint.
\textbf{(3)} \textit{CLIP Similarity (CLIP-\textit{sim}):} The correlation between text and scene content is also assessed using CLIP Similarity. This involves extracting image features via CLIP from rendered frames and calculating the CLIP-Score relative to the text features of the input using the CLIP/H-14 model.
\textbf{(4)} \textit{CLIP Style Similarity (CLIP-style-\textit{sim}):} Similar to CLIP Similarity, this metric focuses on evaluating the correspondence between the scene's overall style and the user's original text input.

We compare \methodname{} against several baselines. \textbf{LayoutGPT+Retrieval} utilizes LayoutGPT\cite{LayoutGPT}, a method that employs a mix of few-shot and zero-shot methods for object placement, with objects subsequently retrieved from the 3D-Front\cite{3DFront} furniture and Objaverse\cite{deitke2023objaverse} databases, retaining their original texture. \textbf{CG+Retrieval} employs our constrained graph-based furniture placement algorithm for layout, followed by object retrieval. \textbf{CG+Inpainting} combines our layout approach with MVDiffusion\cite{tang2023mvdiffusion} for texturing. \methodname{} enriches these with SDS-based layout refinement.

\begin{table}[t]
\centering
\small
\tabcolsep 6pt
\caption{Comparison of layout and content generation quality between \methodname{} and other baselines using Out-of-Boundary Rate (OBB), Caption Similarity (Caption-\textit{sim}), CLIP Similarity (CLIP-\textit{sim}), and CLIP Style Similarity (CLIP-style-\textit{sim}). Mean and variance from 10 runs with random viewpoints are reported. \methodname{} outperforms all baseline methods.}
\begin{tabular}{l|cccc}

\hline
Method            & OOB  & Caption-\textit{sim} & CLIP-\textit{sim} & CLIP-style-\textit{sim} \\ \hline
LayoutGPT+Retrieval & 69.4 & 9.7     & 8.3  & 6.4   \\ 
CG+Retrieval     & 34.2 & 11.2    & 11.5 & 7.0   \\ 
CG+Inpainting   & 34.2 & 15.9    & 27.9 & \textbf{17.5}  \\ 
\methodname{}               & \textbf{23.7} & \textbf{16.2±0.1}    & \textbf{29.3±0.6} & \textbf{17.5±0.5}  \\ \hline
\end{tabular}
\label{tab:compare}
\end{table}
% \vspace{-0.5cm}

\cref{tab:compare} presents the quantitative evaluation. Comparing the first and second rows, using our constrained graph-based representation (CG) significantly improves layout generation over the few-shot LLM-based method without intermediate representations (LayoutGPT), as evidenced by a lower OOB rate and higher content generation scores. Comparing the second and third rows, the egocentric inpainting process further improves text-content and text-style correspondences, elevating these scores (Caption-\textit{sim}, CLIP-\textit{sim}, CLIP-style-\textit{sim}). Comparing the third and fourth rows, the refinement stage further enhances the quality of the layout by reducing overlaps and refining object placements, as indicated by improved scores across all metrics. We include random viewpoint perturbations in the evaluation, such as z-axis rotation and ground translation. The low variances demonstrate the minimal impact of viewpoints on the results.

\subsubsection{Qualitative Comparison for Structure Generation. }
In \cref{fig:result_comparison} panels (a) and (b), we compare our method with existing techniques for floorplan and room layout generation, including zero-shot LLM \cite{openai2023gpt4}, few-shot LLM \cite{LayoutGPT}, and diffusion-based methods \cite{LegoNet}. For house floorplan generation, LLM-based methods often produce rooms with intersecting walls, overly simplistic structures, or illogical configurations, such as a bedroom situated within a bathroom. Our method surpasses these direct LLM-generated plans, especially with abstract prompts, by preserving room relationships and accommodating diverse shapes and sizes. This approach effectively addresses open-vocabulary input scenarios. 
For room layout generation, our SAG-based method demonstrates a clear advantage over traditional few-shot LLMs and diffusion methods. It results in more logically arranged furniture, avoiding common issues such as overlapping bounding boxes and spatial incoherence, particularly in open-vocabulary settings.

\subsubsection{Qualitative Comparison for Texture Inpainting. }
\cref{fig:result_comparison} (c) presents a qualitative analysis of our texture inpainting approach in comparison with other existing methods, guided by the text prompt \texttt{"A gothic living room with wooden furniture."} In this assessment, we evaluate baseline methods such as Stable-Diffusion-2 with depth conditioning (SD images) and the results of back-propagating image pixel colors to mesh vertices using differentiable rendering (SD texture). Additionally, we examine images generated by MVDiffusion (MVD images) and their subsequent mesh vertex back-propagation (MVD texture). While SD images exhibit richly detailed textures, they often lead to inconsistencies across different views, which undermines the effectiveness of back-propagation. In contrast, MVD images demonstrate a higher level of view consistency. The back-propagation of MVD images (MVD texture) further enhances this consistency across various viewpoints. By integrating this method, which leverages the strengths of MVDiffusion and differentiable rendering, we achieve textures that are not only consistent but also smoothly transitioned, accurately reflecting the textual prompt provided.

% \vspace{-2cm}
\section{Conclusion}
In this paper, we present \methodname{}, a novel framework designed to generate house-scale scenes from open-vocabulary textual inputs. Leveraging specific prompt templates, our system effectively interprets, follows, and enriches user-provided text to create detailed scenes. Utilizing a two-stage generation approach, which combines LLM-designated rules with SDS loss refinement, our method proficiently arranges a diverse array of objects in a realistic manner. The resulting houses feature a structured geometry representation, enhancing the ease of user editing and modification. We believe that \methodname{} paves the way for a wide range of applications, including complex 3D interior design, immersive augmented and virtual reality experiences, dynamic gaming environments, and advanced training modules for embodied agents.

\noindent \textbf{Limitations and Future Work.}
Still, \methodname{} faces limitations due to the current LLMs' understanding of 3D spaces and the challenges of maintaining consistency in multi-view inpainting using existing techniques. Future work will focus on overcoming these challenges to improve the system's capacity for generating detailed and coherent 3D environments from textual inputs. 
\section*{Acknowledgements} This research was supported by AFOSR grant FA9550-21-1-0214. The authors thank Dylan Hu, Selena Ling, Kai Wang and Daniel Ritchie.

% ---- Bibliography ----
%
% BibTeX users should specify bibliography style 'splncs04'.
% References will then be sorted and formatted in the correct style.
%
\bibliographystyle{splncs04}
\bibliography{main}

\newpage 
\begin{center}
   \Large \textbf{Supplementary Material for \methodname{}}
\end{center}

% This is the Supplementary Material for \methodname{}. It includes additional visualizations and experimental results, featuring trajectory generation visualizations (Sec.~\ref{sec:trajectory}), layout refinement visualizations (Sec.~\ref{sec:sds}), and a qualitative and quantitative comparison with contemporary work (Sec.~\ref{sec:compare}), highlighting our method's uniqueness. We also analyze the limitations of the current method (Sec.~\ref{sec:limitation}). Furthermore, we provide details on our method's settings, including implementation details (Sec.~\ref{sec:details}), utilized prompts (Sec.~\ref{sec:prompts}), and the rules and algorithms for object placement (Sec.~\ref{sec:rules}).

This is the Supplementary Material for \methodname{}. It includes additional visualizations and experimental results, featuring video visualizations that navigate through the interiors of generated houses (Sec.~\ref{sec:videos}), trajectory generation visualizations (Sec.~\ref{sec:trajectory}), layout refinement visualizations (Sec.~\ref{sec:sds}), complex texture generation (Sec.~\ref{sec:texture}), diverse floorplans (Sec.~\ref{sec:floorplan}), and a qualitative and quantitative comparison with contemporary work (Sec.~\ref{sec:compare}), highlighting our method's uniqueness. We present the time consumption breakdown for each component (Sec.~\ref{sec:time}). We also analyze the limitations of the current method (Sec.~\ref{sec:limitation}). Furthermore, we provide details on our method's settings, including implementation details (Sec.~\ref{sec:details}), utilized prompts (Sec.~\ref{sec:prompts}), and the rules and algorithms for object placement (Sec.~\ref{sec:rules}).
\section{House Egocentric Visualizations with Videos}
\label{sec:videos}
\begin{figure}[h]
    \centering
    \begin{subfigure}[b]{0.45\textwidth}
        \includegraphics[width=0.98\textwidth]{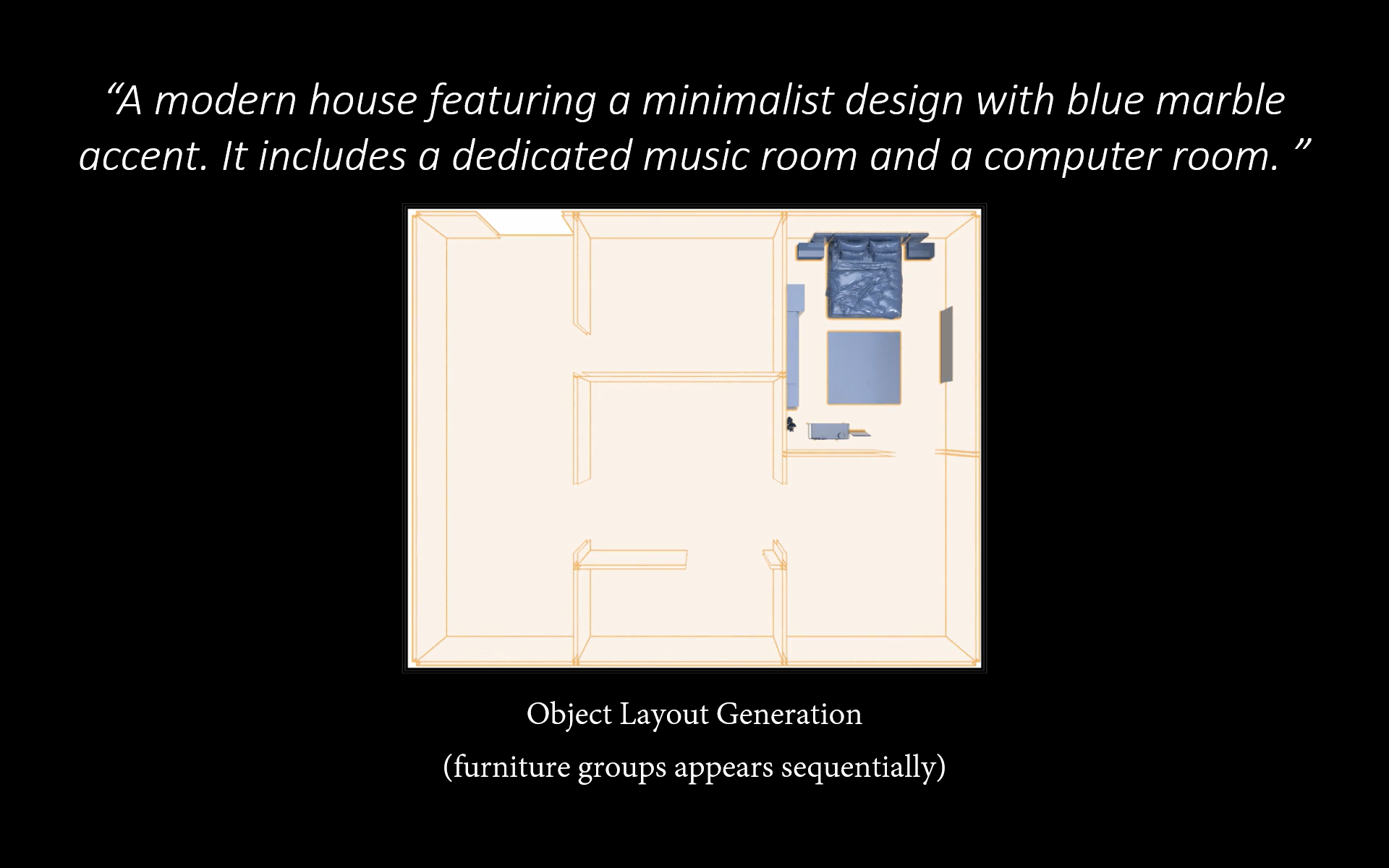} % Replace with your image path
        \caption{Layout Generation Visualization.}
        \label{fig:vidoe1}
    \end{subfigure}
    % \hfill
    \begin{subfigure}[b]{0.45\textwidth}
        \includegraphics[width=0.98\textwidth]{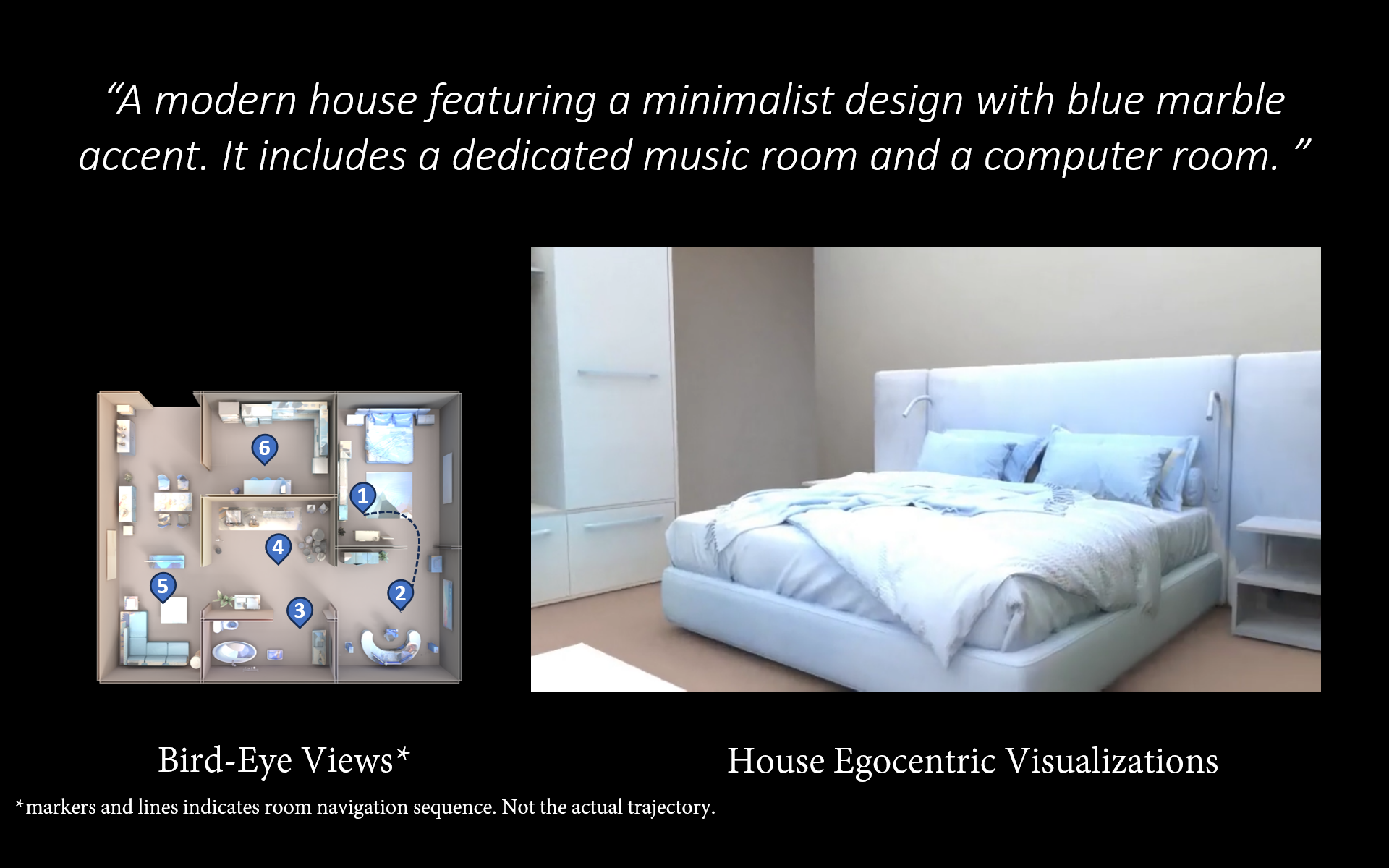} % Replace with your image path
        \caption{House Egocentric Tour Visualization.}
        \label{fig:vidoe2}
    \end{subfigure}
    % \vspace{-0.2cm}
    \caption{Supplementary video provide layout generation visualization and house egocentric visualizations.}
    \label{fig:vidoes}
\end{figure}

A Supplementary Video is included to offer additional qualitative results. Fig.~\ref{fig:vidoes} presents screenshots taken from the Supplementary Video. This video demonstrates the sequential placement of furniture groups within the layout generation process (see Fig.~\ref{fig:vidoe1}), effectively showcasing the nuances of our placement algorithm (refer to Algorithm~\ref{alg:layout}). Furthermore, it features egocentric navigation through the generated rooms (see Fig.~\ref{fig:vidoe2}), highlighting how our methods for egocentric refinement and in-painting achieve coherent layouts and create realistic textures across different viewpoints.

\newpage
\section{Egocentric Trajectory Generation}
\label{sec:trajectory}
\begin{figure}[ht]
\begin{center}
\includegraphics[width=0.98\textwidth]{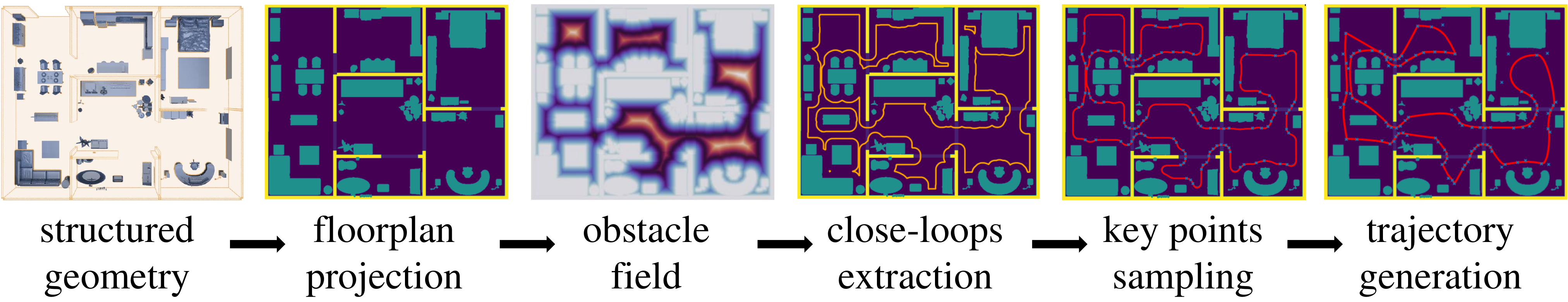}
\end{center}
\caption{\textbf{Egocentric Trajectory Generation Pipeline.} Given the base mesh generated from amodal representations, floorplan projection, obstacle field and close-loops are extracted sequentially. Key points are then sampled and interpolated to obtain the final trajectory. }
\label{fig:trajectory}
\end{figure}

Fig.~\ref{fig:trajectory} delineates the process of egocentric trajectory generation, an essential element that fosters object-level layout refinement and texture in-painting. Starting with the structured geometry derived from amodal representations, we project this geometry onto a 2D plane to obtain a floor plan. From this floor plan, we extract an obstacle field, which calculates the nearest distance from each point to furniture or walls. Utilizing these distance fields, we identify closed loops that are subsequently vectorized for refinement. This refinement process includes the elimination of short loops and areas of high curvature, along with the application of Gaussian smoothing to address minor irregularities. To facilitate visibility of objects and ensure navigation through doorways during exploration, keypoints are meticulously placed around these critical features on the closed loops. The final trajectory is constructed by interpolating between these strategically sampled keypoints, ensuring a coherent path that enhances the scene's navigability and visibility.

\newpage
\section{Layout Refinement Visualization}
\label{sec:sds}
\begin{figure}[ht]
\begin{center}
\includegraphics[width=\textwidth]{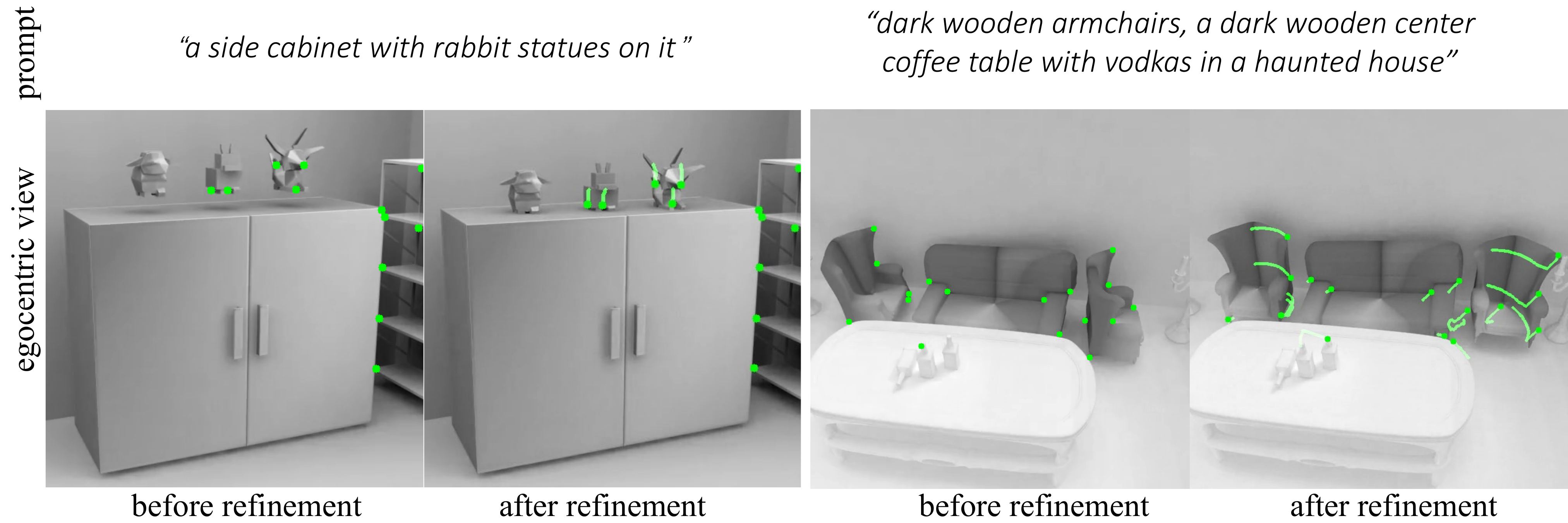}
\end{center}
\vspace{-0.5cm}
\caption{\textbf{Layout Refinement Visualization.} The refinement trajectory is visualized in green color through optical flow. Scene layout shows enhanced alignment with the textual prompt following layout refinement.}
\label{fig:result_refine}
\end{figure}

Fig.~\ref{fig:result_refine} visualizes object layouts before and after refinement. The green highlighted lines illustrate the objects' positions throughout the optimization process from a single viewpoint. The key-points are extracted by Scale-Invariant Feature Transform (SIFT) and tracked by optical flow. The results indicate that the refinement process effectively anchors floating items (left) and improves object orientation while correcting asymmetry (right).
\newpage
\section{Complex Texture}
\label{sec:texture}
\begin{figure}[h]
    \centering
    \includegraphics[width=0.80\textwidth]{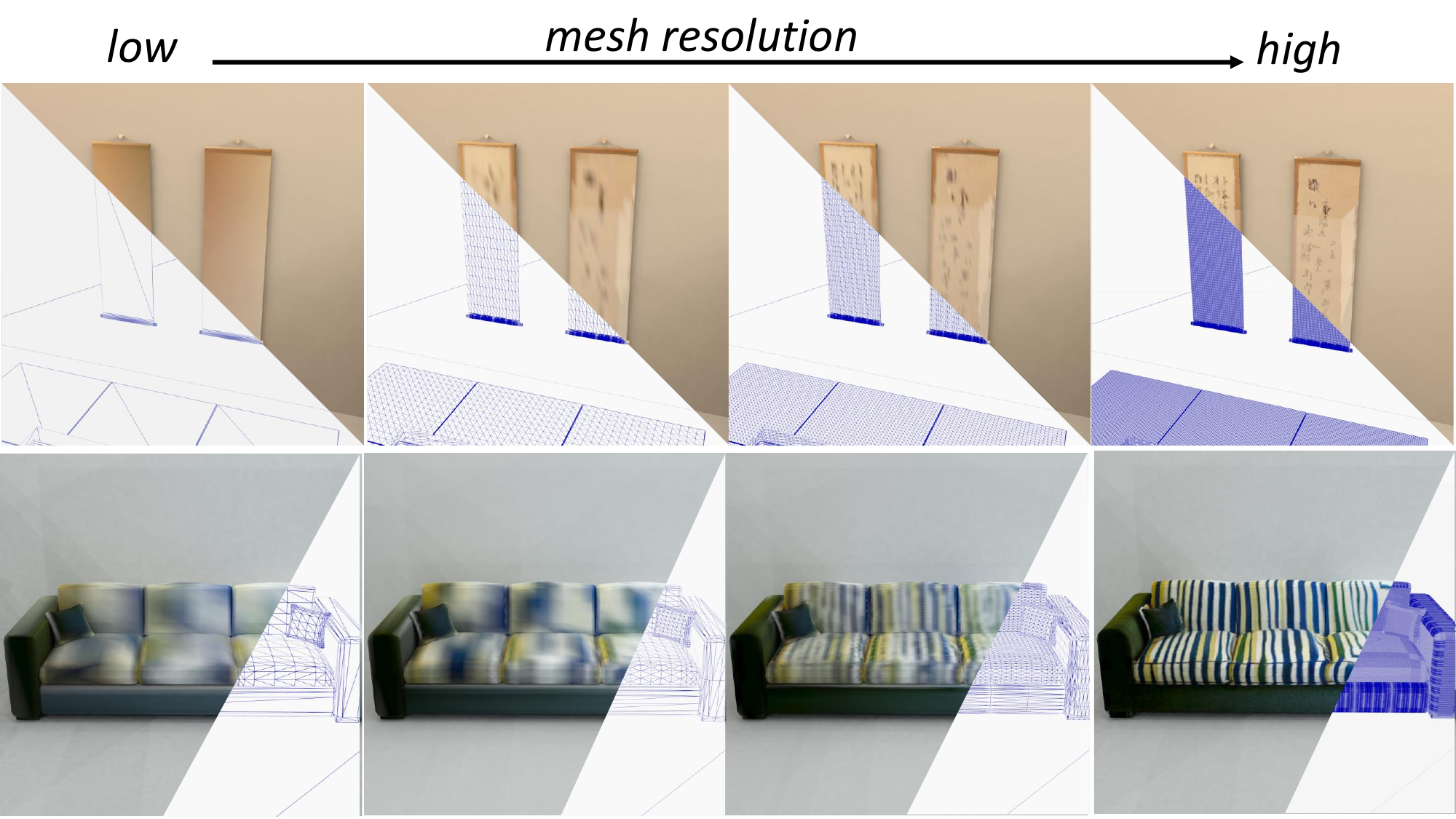}
    \caption{\textbf{Generating Complex Texture.} AnyHome is able to generate complex texture as mesh resolution increases from left to right.}
    \label{fig:texture}
\end{figure}

Fig.~\ref{fig:texture} shows Anyhome's ability to generate both complex details (the calligraphy) and decorative patterns (the stripe). The complexity of the texture is limited by the mesh resolution of object datasets (Objaverse and 3D Future)—complex patterns cannot be applied to a surface consisting of only two triangular meshes. Instead, we can re-mesh these objects in the original dataset to a higher resolution to accommodate more complex textures, albeit at the cost of increased inference time. As shown in the figure, the clarity of both the calligraphy and stripe patterns improves with higher mesh resolution.

\newpage
\section{Floorplan Diversity}
\label{sec:floorplan}
\begin{figure}[h]
    \centering
    \includegraphics[width=0.80\textwidth]{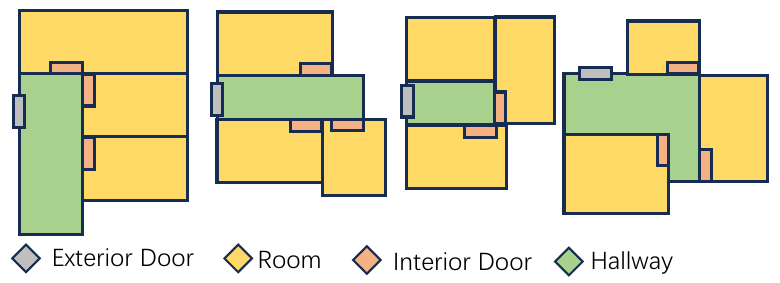}
    \caption{\textbf{Generating Novel Room-Type.} Given the prompt \texttt{"A 1B1B apartment with a hallway."}, AnyHome is able to generate novel room types (i.e., the hallway) and diverse floorplans.}
    \label{fig:hallway}
\end{figure}

\cref{fig:hallway} shows that AnyHome can generate diverse floorplans given the same prompt \texttt{"A 1B1B apartment
with a hallway."}. The generation of novel room types is made possible by the LLM’s ability of mapping unconventional room type (hallway) to types that exist in the model’s training set (i.e., entrance). At the same time, the diversity results from HouseGAN++'s generalizability. 

\section{Time Consumption}
\label{sec:time}
\begin{figure}[h]
    \centering
    \includegraphics[width=0.80\textwidth]{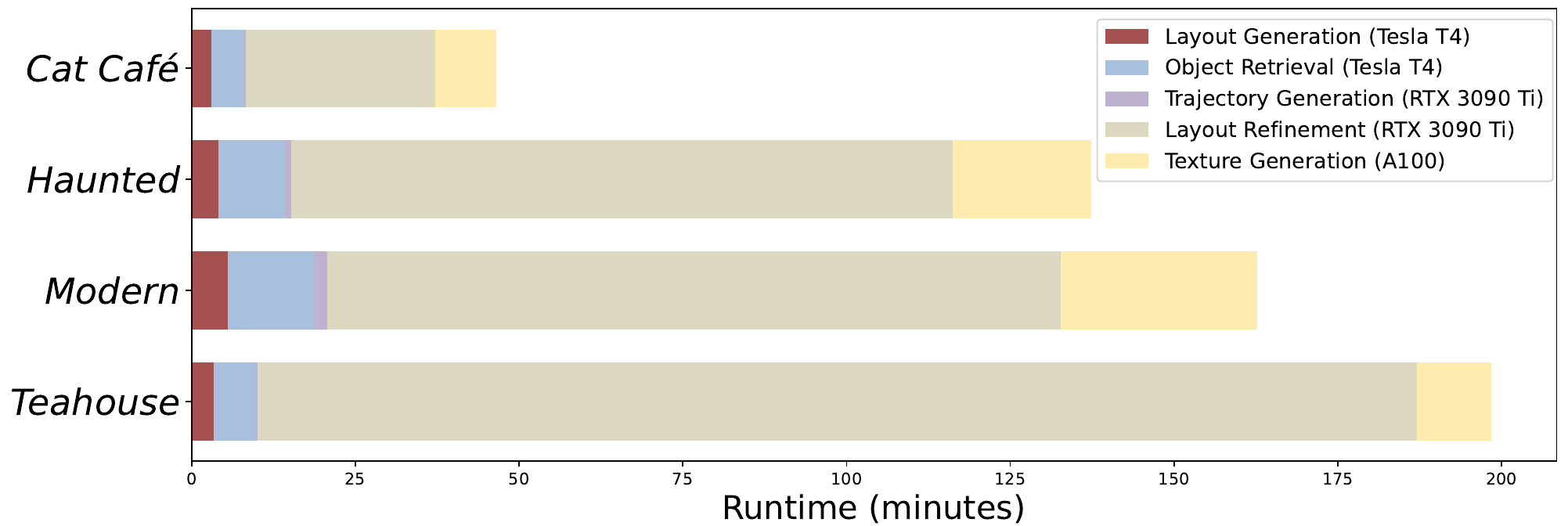}
    \caption{\textbf{Time Consumption.} The time consumption of each component for examples shown in the main paper.}
    \label{fig:time}
\end{figure}

\cref{fig:time} presents a detailed breakdown of time consumption for the four sample houses included in the main paper.

\clearpage
\section{Comparison with Contemporary Works}
\label{sec:compare}
\begin{figure}[ht]
\begin{center}
\includegraphics[width=0.98\textwidth]{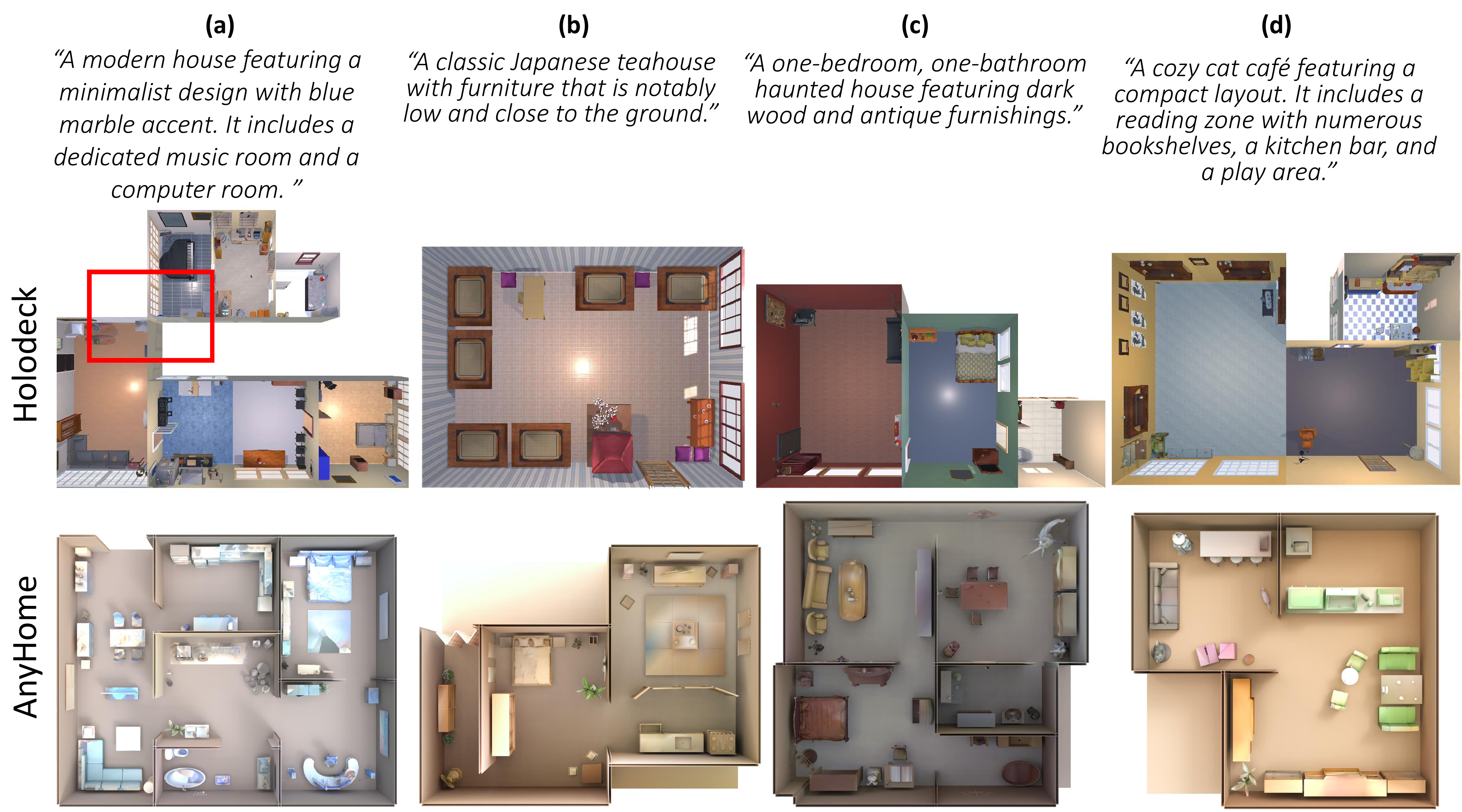}
\end{center}
\caption{\textbf{Qualitative Comparison with Holodeck.} 
\methodname{} proves its uniqueness and competitiveness through its ability to generate more logical and complex floorplans, achieve a broader array of layout designs, and provide consistent and customizable texture creation when compared to contemporary works such as Holodeck.}
\label{fig:holodeck}
\end{figure}

\noindent Holodeck\cite{yang2024holodeck} represents a contemporary method for text-conditioned indoor scene generation. Although direct comparisons with contemporaneous works are not obligatory, our goal is to highlight \methodname{}'s unique attributes and competitive edge. Fig.~\ref{fig:holodeck} offers a qualitative comparison with Holodeck, illustrating differences in several key aspects.

\textbf{Floorplan Generation.} Holodeck is capable of generating regular floorplans directly through LLMs. However, it often produces impractical floorplans for complex scenes, as highlighted in red in panel \textbf{a}, and the floorplans tend to be overly simplistic, as observed in panels \textbf{b} and \textbf{c}. In contrast, \methodname{} generates realistic and diverse floorplans, featuring various functional rooms that enhance the overall functionality of the house.

\textbf{Room Object Selection.} While Holodeck selects appropriate furniture for common room types, such as a bed for a bedroom in panel \textbf{c}, it struggles to identify suitable objects for less common room types, leading to empty spaces like the cat café room in panel \textbf{d}. It also shows limitations in the quantity and diversity of objects, evidenced by the sparse furnishings in panels \textbf{c} and \textbf{d} and repetitive furniture in panel \textbf{b}. In contrast, \methodname{} adeptly selects furniture corresponding to both the room type and the house's description, resulting in efficiently utilized and densely furnished spaces.

\textbf{Room Layout Generation.} Holodeck tends to position objects along walls, resulting in less varied layouts. Conversely, \methodname{} facilitates more realistic and varied room layouts, thanks to placement rules informed by Large Language Models (LLMs).

\textbf{Object Texture.} Objects in Holodeck’s scenes exhibit a wide range of textures, inherently tied to the object meshes, leaving limited room for customization. On the contrary, \methodname{} applies texture to object meshes through diffusion models with style-specific prompts, enabling texture customization and visual domain adaptation. This approach produces scenes with enhanced texture consistency and better alignment with the input description, exemplified by the creation of a "blue marble accent" in panel \textbf{a}.

\begin{table}[t]
\centering
\small
\tabcolsep 6pt
\caption{Quantitative comparison between Holodeck and \methodname{} using GPT-4V.}
\begin{tabular}{l|ccccc}
\hline
Method            & \textit{Prompt-Align}  & \textit{Layout} & \textit{Object} & \textit{Texture} & \textit{Overall} \\ \hline
Holodeck\cite{yang2024holodeck} & 7.8 & 6.1 & 4.3 & 5.8 & 6.0   \\ 
\methodname{}    & \textbf{9.0} & \textbf{8.7} & \textbf{7.8} & \textbf{6.0} & \textbf{7.8}  \\  \hline
\end{tabular}
\label{tab:holodeck}
\end{table}

\begin{figure}[ht]
\begin{center}
\includegraphics[width=0.98\textwidth]{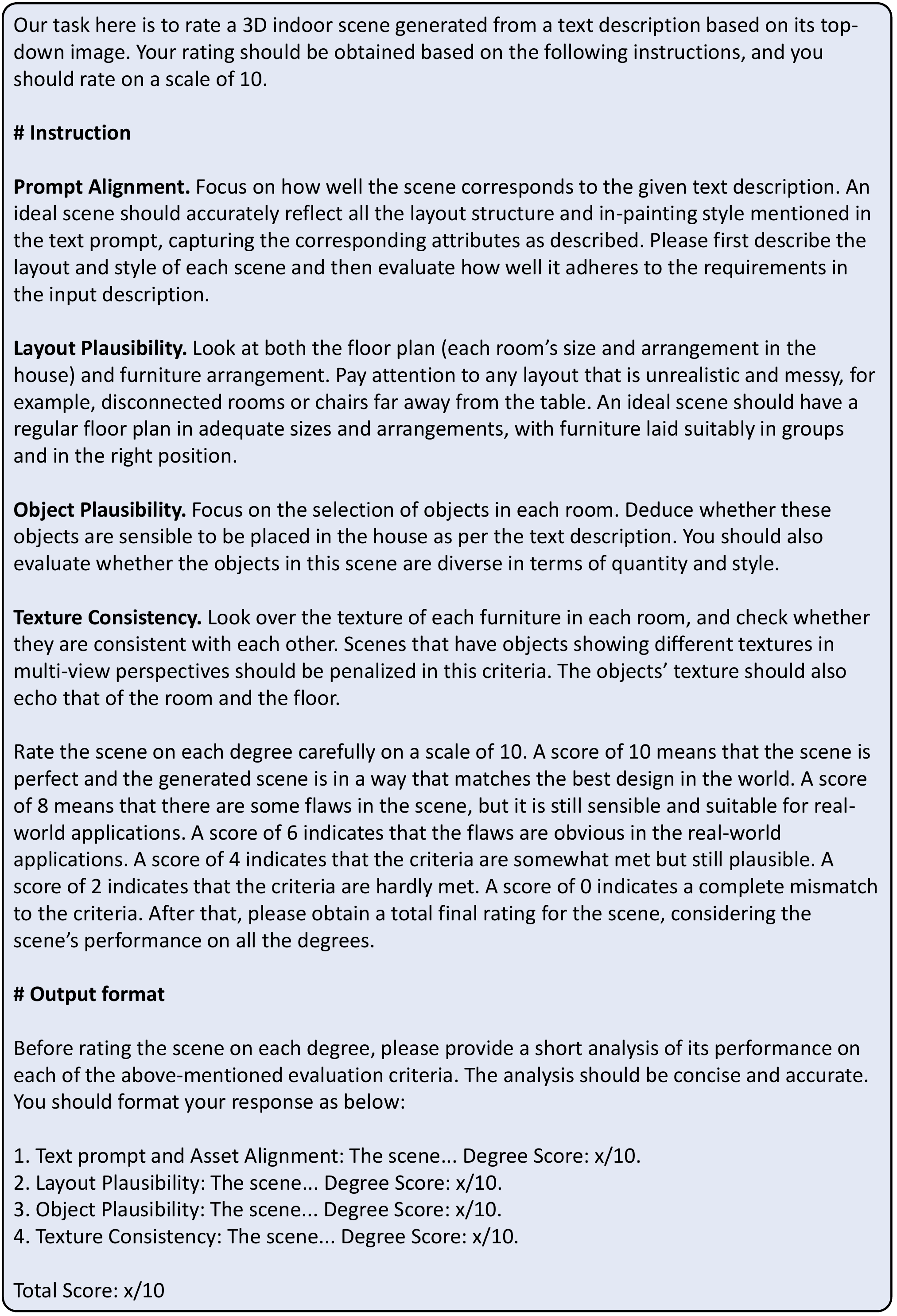}
\end{center}
\caption{Prompts for GPT-V4 evaluation.}
\label{fig:gpt4v}
\end{figure}

As Holodeck\cite{yang2024holodeck} does not support egocentric views or prompts, a direct comparison with \methodname{} using the metrics outlined in the main paper is not feasible. Consequently, following the approach of \cite{wu2024gpt}, we employ GPT-4V for evaluation by providing it with top-down views of the generated scenes. Table~\ref{tab:holodeck} compares our method with Holodeck across various dimensions: the scene’s alignment with the given prompt (\textit{Prompt-Align}), the realism of house plans and room layouts (\textit{Layout}), the authenticity of objects (\textit{Object}), and the consistency of textures (\textit{Texture}). The specific prompts used for these evaluations and the scoring scale are detailed in Fig.~\ref{fig:gpt4v}.

Quantitative analysis reveals that \methodname{} excels in aligning scenes with the provided prompts, achieving an impressive average score of \texttt{9.0} out of \texttt{10} for \textit{Prompt-Align}. This superiority stems from its capability to generate realistic floorplans for any designated room types, accurately retrieve objects corresponding to a room’s function and style, and apply texture in line with the prompt’s specifications. In individual assessments, \methodname{} also stands out. Its two-stage approach for floorplan and room layout generation yields average scores of \texttt{8.7} for \textit{Layout} and \texttt{7.8} for \textit{Object}, demonstrating the generated geometry's plausibility and the robustness of this generation. Notably, Holodeck scores relatively low in \textit{Object}, primarily due to its tendency to create sparsely furnished rooms, as depicted in Fig.~\ref{fig:holodeck}. Both \methodname{} and Holodeck show limitations in achieving texture consistency (\textit{Texture}), with \methodname{} occasionally painting a single object in inconsistent colors, as illustrated in Fig.~\ref{fig:failure}. This issue arises from inconsistencies in viewpoint and the images generated from the multi-view diffusion inpainting model\cite{tang2023mvdiffusion}. Overall, \methodname{} attains a score of \texttt{7.8}, indicating the presence of flaws but affirming the method's practical applicability in real-world scenarios.

\clearpage
\newpage
\section{Limitations and Failure Cases}
\label{sec:limitation}
\begin{figure}[ht]
\begin{center}
\includegraphics[width=0.98\textwidth]{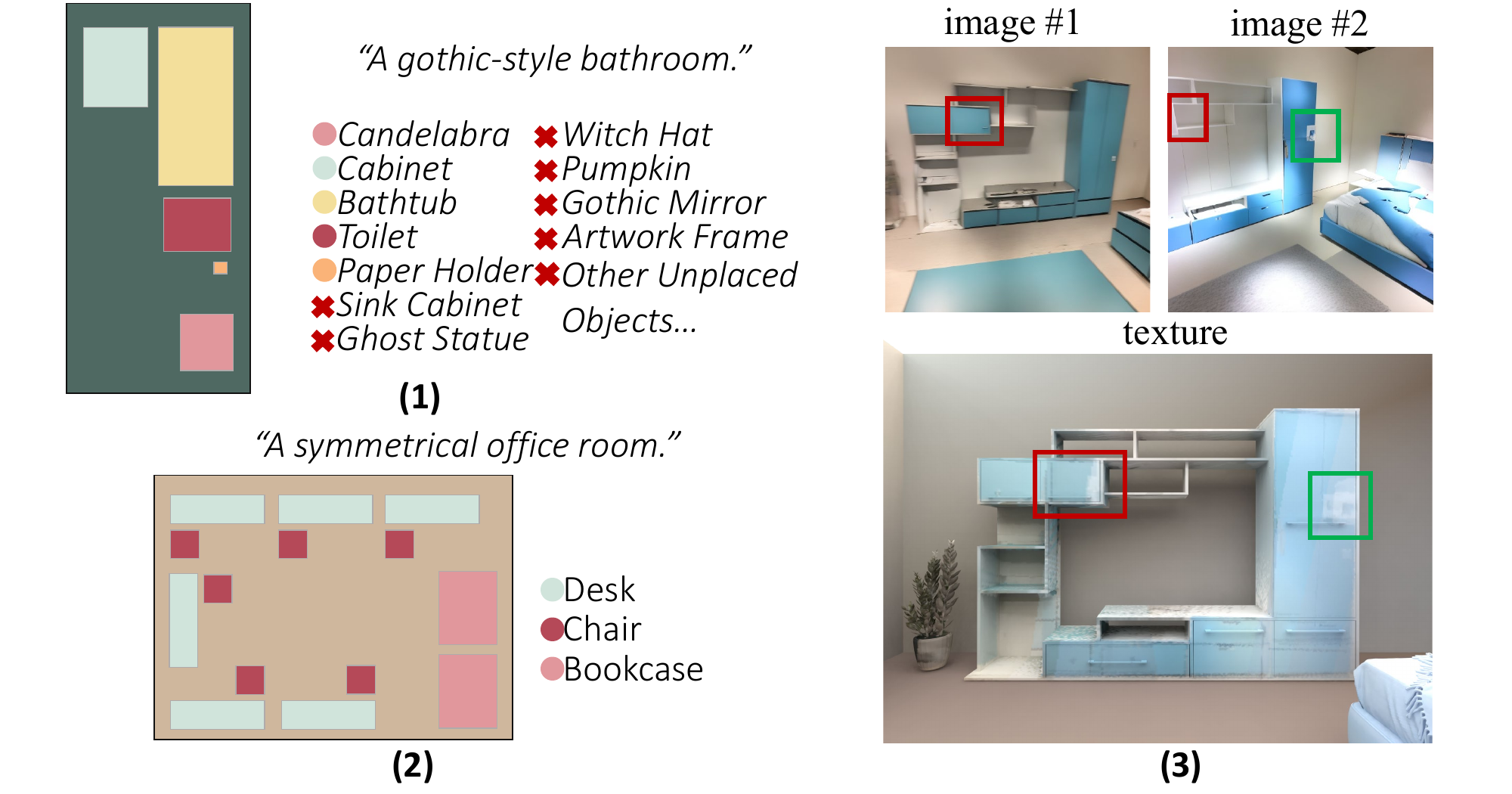}
\end{center}
\caption{\textbf{Limitations.} (1) Generation of overly extensive lists by the LLM for confined spaces. (2) Inability to adhere to particular layout constraints stemming from the LLM's interpretation of the rules. (3) Asymmetrical textures arising from inconsistencies in input images sourced from various viewpoints.}
\label{fig:failure}
\end{figure}

\methodname{}, as a pioneering approach to generating customized, house-scale scenes, faces several challenges. First, despite repeatedly emphasizing room dimensions to the Large Language Model (LLM), it occasionally places too many objects in limited spaces (see Fig.~\ref{fig:failure} \textbf{(1)}), highlighting its limitations in grasping sizes and spatial relationships. Although the placement algorithm stops adding objects once space is exhausted, ensuring the final layout remains viable, the initial list of objects and the suggested placement rules from the LLM do not always result in a logically arranged layout. Second, while \methodname{} includes rules capable of generating symmetrical layouts—such as classrooms with tables and chairs in orderly rows and columns, as shown in Fig. 7 of the main paper—the LLM sometimes misunderstands these rules. It selects rules that place furniture reasonably but fails to comply with the style specified in the prompt, as evident in (Fig.\ref{fig:failure} \textbf{(2)}), where desks and chairs are realistically but incorrectly arranged around the walls. Lastly, our egocentric inpainting approach, though it facilitates realistic texture customization, encounters issues with view consistency. Fig.~\ref{fig:failure} \textbf{(3)} illustrates these challenges, where red boxes highlight asymmetrical textures resulting from the disparate images generated by the multi-view diffusion inpainting model \cite{tang2023mvdiffusion} across different viewpoints, and green boxes show the loss of texture detail due to viewpoint inconsistencies.

\clearpage
\newpage
\section{Implementation Details}\label{sec:details}

\noindent\textbf{LLM-guided Layout Generation.} We utilize OpenAI's \texttt{gpt-4-1106-preview} model as the LLM for our system and set the \texttt{temperature} to \texttt{0.7} to allow diversity in each generation. Simultaneously, we set \texttt{Top-p} to \texttt{1} and apply no penalty to the next token prediction. 

\noindent\textbf{Diffusion-guided Refinement.} For layout refinement, we employ Stable Diffusion 2 as our diffusion model. In the diffusion process, we select a batch of 8 images representing multiple views, all associated with the same prompt, for optimization. Beyond using normal maps paired with diffusion prompts, we also introduce a technique where objects are rendered in random colors—such as "red," "blue," "green," "yellow," "purple," and "orange"—and these images are paired with prompts modified to reflect these colors for enhanced refinement. For instance, if rabbits are rendered in blue and a cabinet in red, the original egocentric diffusion prompt "a side cabinet with rabbit statues on it" is adjusted to "a red side cabinet with blue rabbit statues on it." This color-modified approach allows for more precise refinement by providing the diffusion model with additional context about the scene's color composition, thus improving the alignment between the generated images and the specified layout and appearance.

\noindent\textbf{Diffusion-guided Inpainting.} We use MVDiffusion as our multi-view, depth-conditioned inpainting model\cite{tang2023mvdiffusion}, setting a \texttt{guidance\_scale} of \texttt{15} to increase the influence of text prompts in generation and an \texttt{overlap\_filter} of \texttt{0.1} to ensure maximal consistency.

\section{Prompts}
\label{sec:prompts}
\begin{figure}[ht]
\begin{center}
\includegraphics[width=0.98\textwidth]{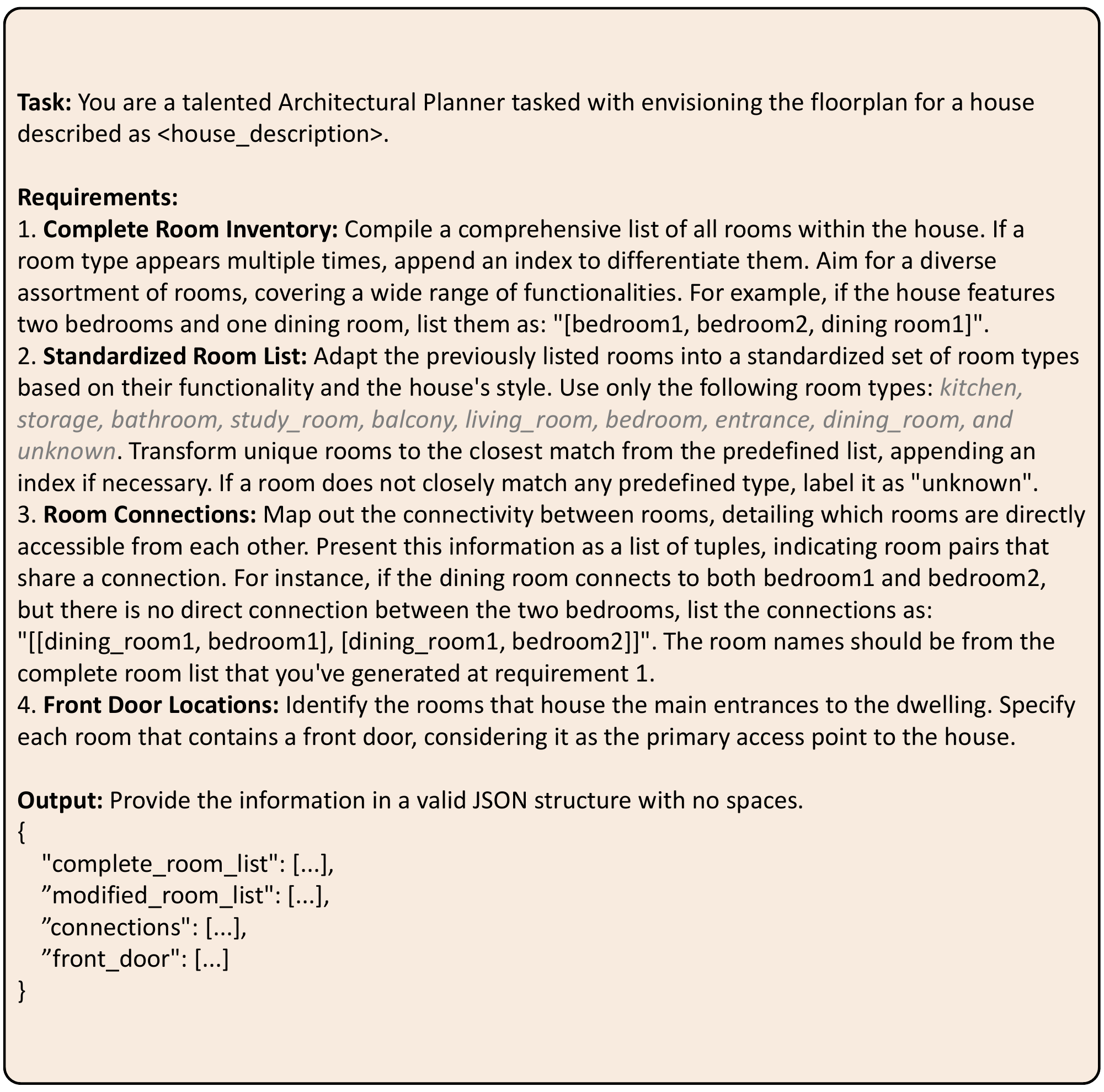}
\end{center}
\caption{$p_{\text{floorplan}}$ and $p_{\text{map}}$ for floor plan generation.}
\label{fig:promtp1}
\end{figure}

\begin{figure}[ht]
\begin{center}
\includegraphics[width=0.98\textwidth]{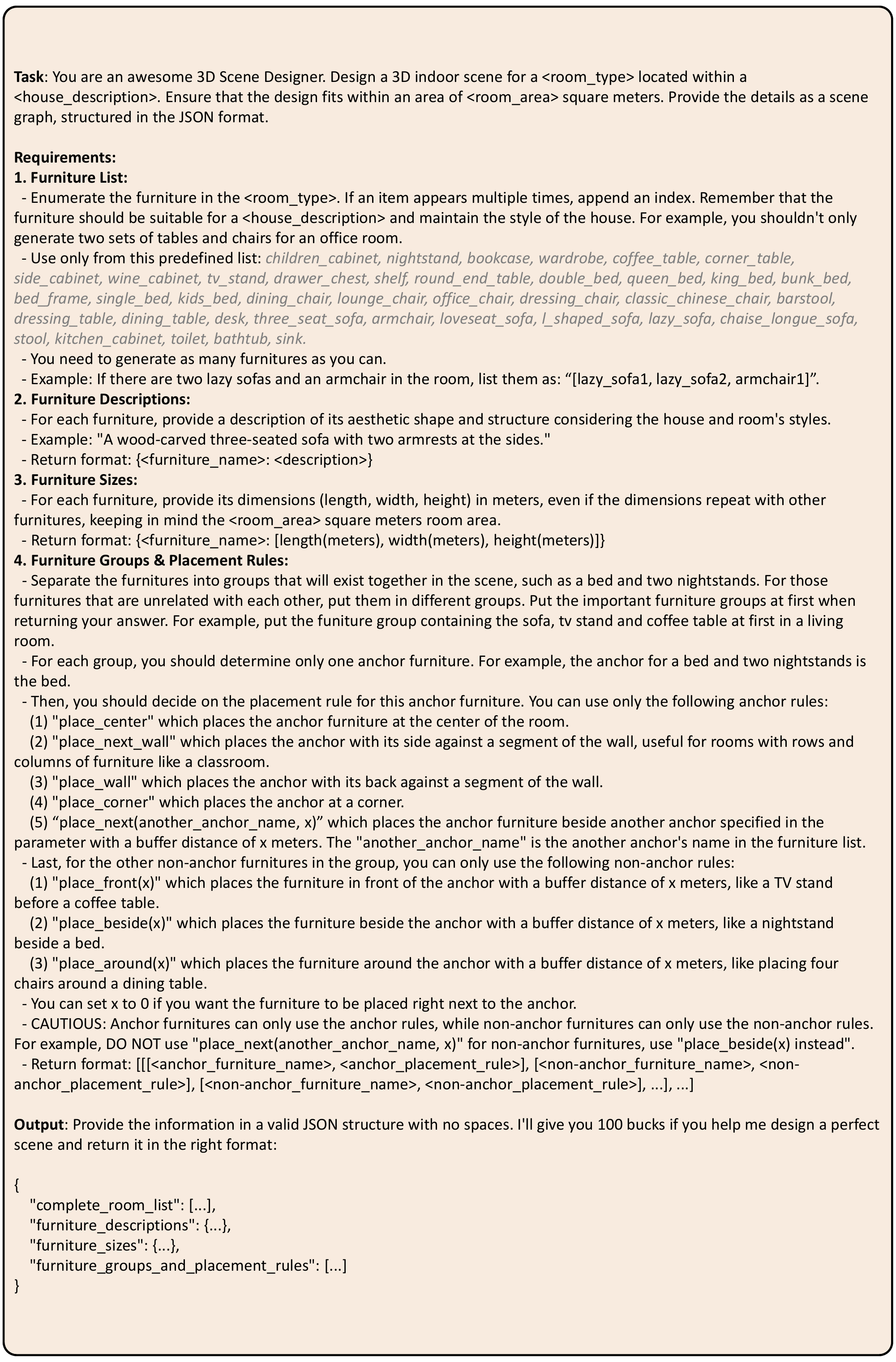}
\end{center}
\caption{$p_{\text{room}}$ for room layout generation - the $p_{\text{furniture}}$ part.}
\label{fig:promtp2}
\end{figure}

\begin{figure}[ht]
\begin{center}
\includegraphics[width=0.98\textwidth]{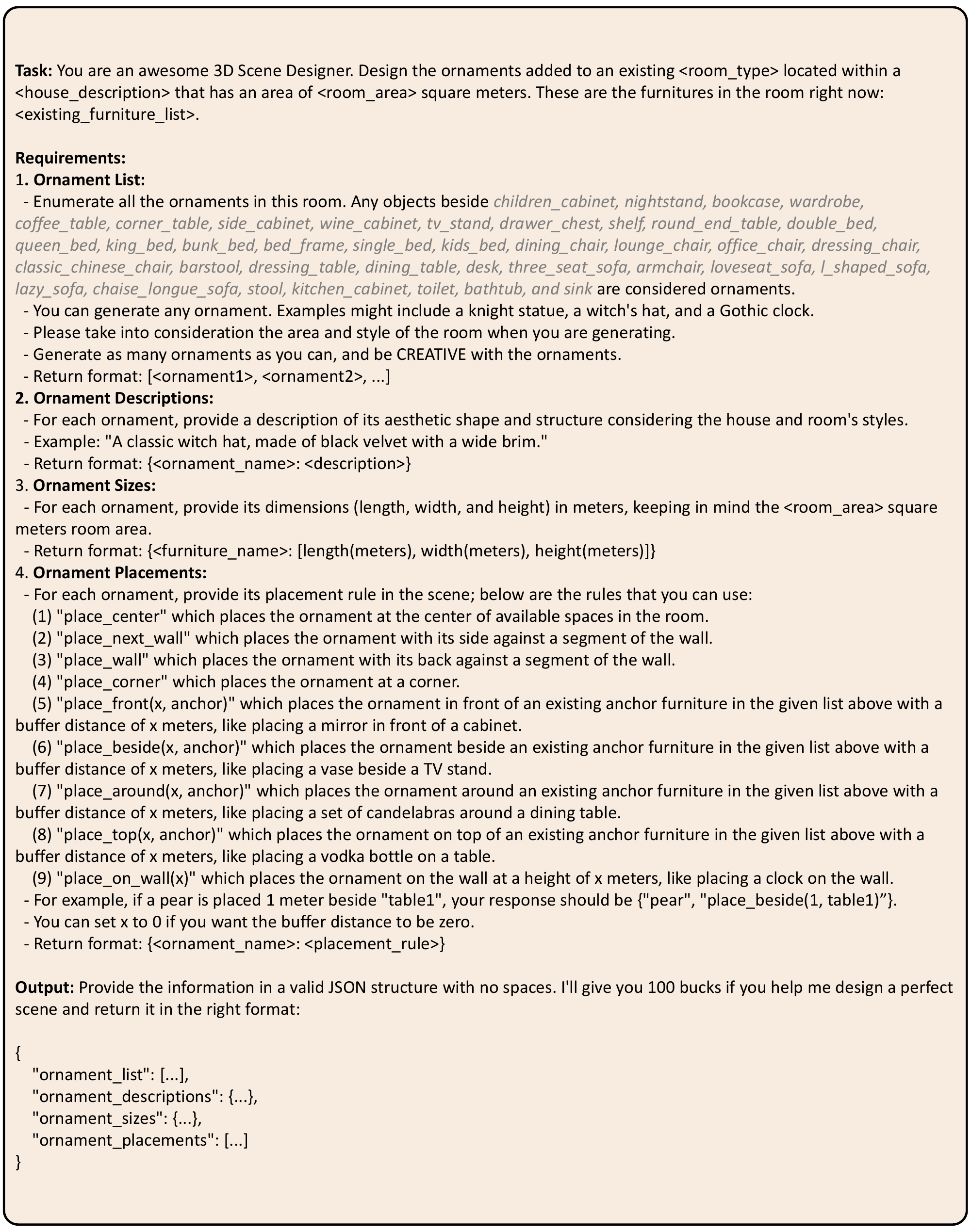}
\end{center}
\caption{$p_{\text{room}}$ for room layout generation - the $p_{\text{ornament}}$ part.}
\label{fig:promtp3}
\end{figure}

\begin{figure}[ht]
\begin{center}
\includegraphics[width=0.98\textwidth]{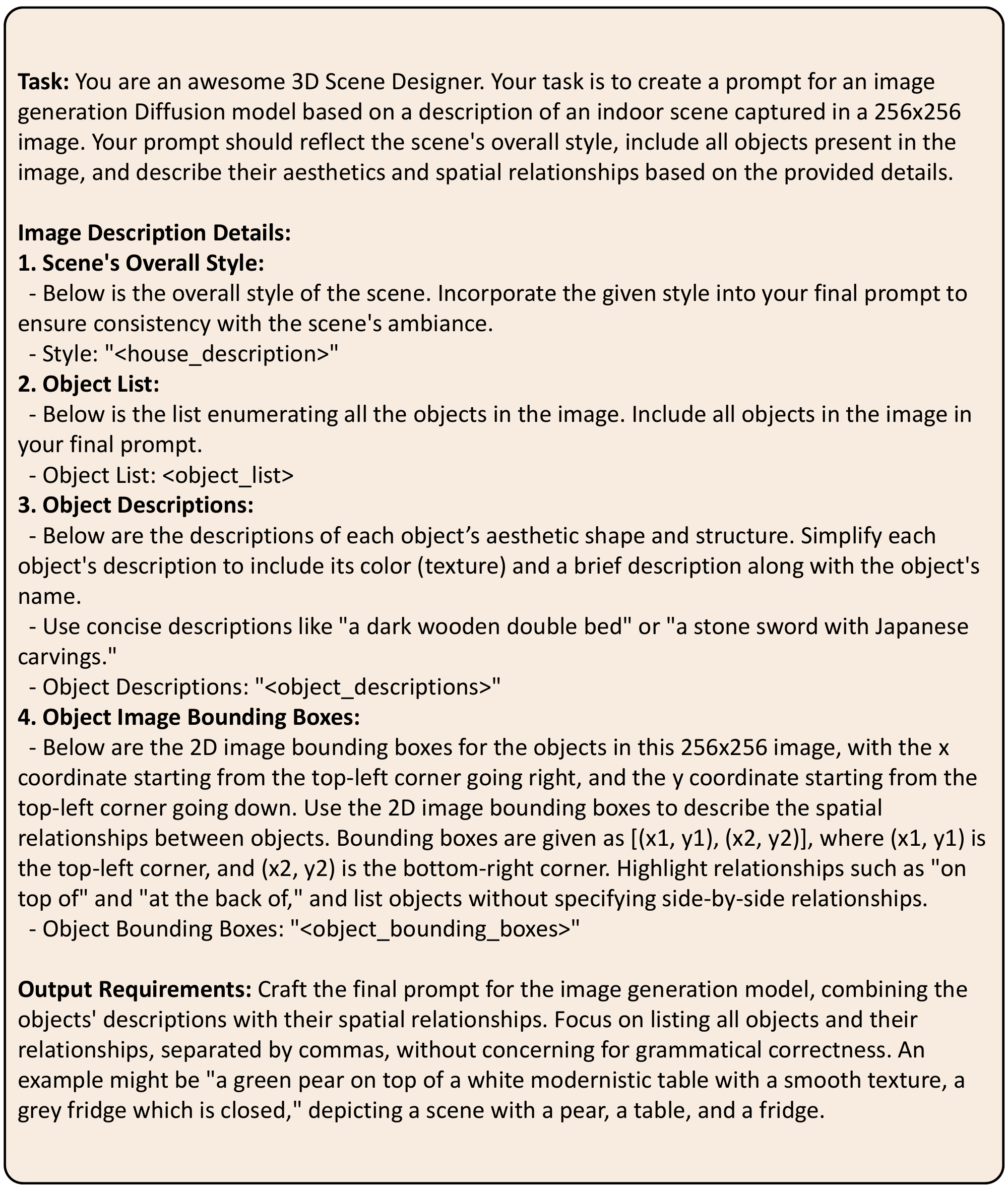}
\end{center}
\caption{$p_{\text{appearance}}$ for in-painting prompt generation.}
\label{fig:promtp4}
\end{figure}

\begin{figure}[ht]
\begin{center}
\includegraphics[width=0.98\textwidth]{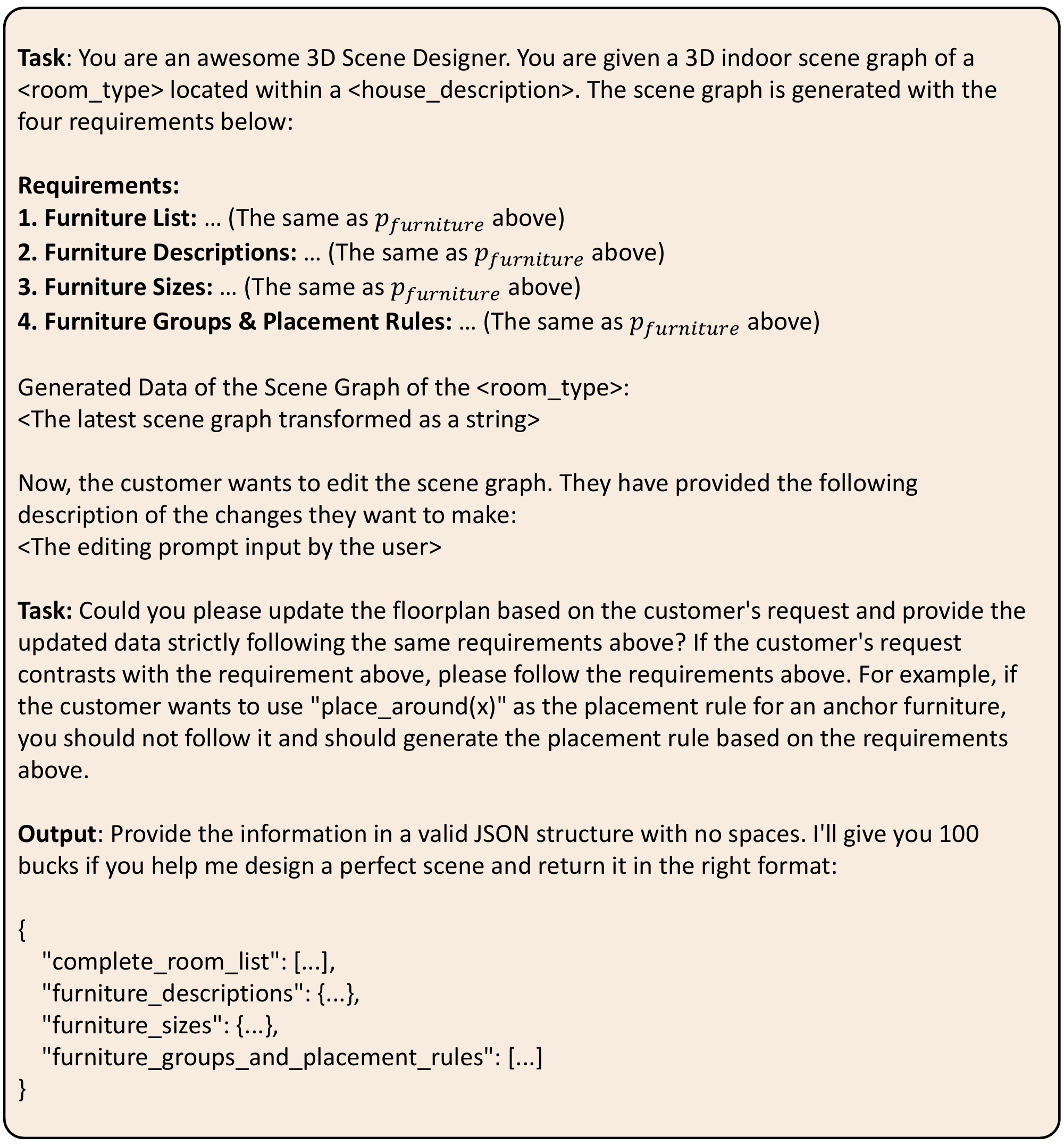}
\end{center}
\caption{$p_{\text{furniture\_edit}}$ editing furniture-related properties.}
\label{fig:promtp5}
\end{figure}

\begin{figure}[ht]
\begin{center}
\includegraphics[width=0.98\textwidth]{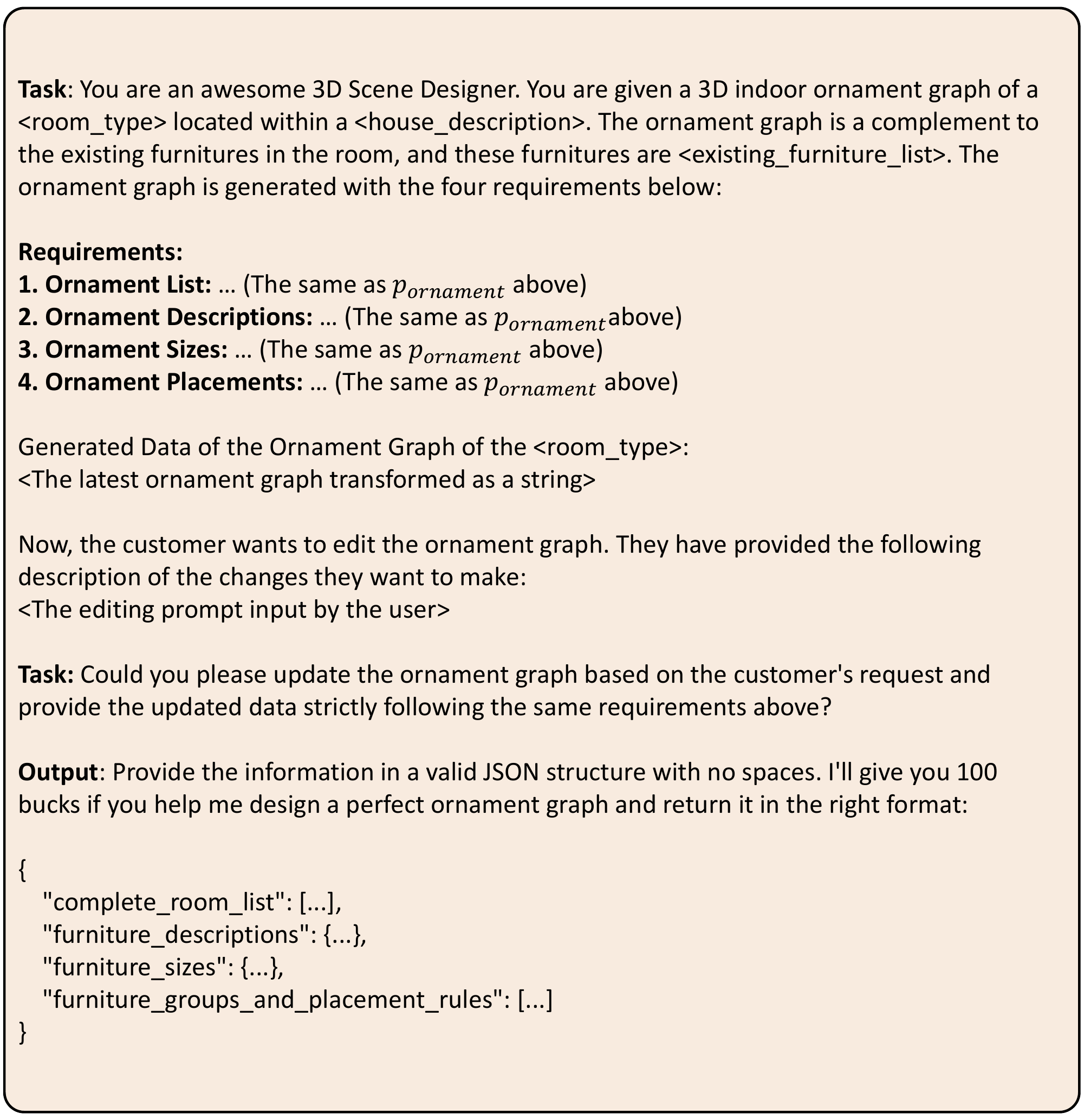}
\end{center}
\caption{$p_{\text{ornament\_edit}}$ for ornament-related editing.}
\label{fig:promtp6}
\end{figure}

Fig.~\ref{fig:promtp1}, \ref{fig:promtp2}, \ref{fig:promtp3}, \ref{fig:promtp4}, \ref{fig:promtp5} and \ref{fig:promtp6} present the complete prompt templates employed by \methodname{}. The prompt $p_{\text{floorplan}}$ is designed to generate a bubble diagram for house floorplans, while $p_{\text{map}}$ translates unconventional room types into those included in the RPLAN dataset\cite{RPLAN} to facilitate floorplan generation. These prompts are combined for efficient generation. The $p_{\text{room}}$ prompt produces a room constraint graph, detailing the description, dimensions, and placement of objects. It is divided into two distinct prompts, $p_{\text{furniture}}$ and $p_{\text{ornament}}$ for furniture and ornaments respectively, to address the Large Language Model (LLM)'s limitations in generating diverse scenes with complex descriptions in a single prompt round. $p_{\text{furniture\_edit}}$ and $p_{\text{ornamen\_edit}}$ are the corresponding prompts for editing. The $p_{\text{appearance}}$ prompt directs the LLM to provide inpainting prompt for each depth map by describing the objects and their image bounding boxes, which are calculated from viewpoints sampled by the egocentric trajectory.

Our prompt templates are designed to: (1) engage the LLM as an assistant for 3D scene generation, (2) outline requirements systematically, and (3) specify the desired output format, specifically requesting responses in JSON to streamline parsing and reduce irrelevant information. To improve adherence to expected outputs, we include task examples for the LLM. Notably, to address instances where LLMs overlook specified room types or styles—such as placing a double bed in a living room or choosing modern furniture for a medieval castle—we reinforce these details in the prompts to enhance consistency with given inputs.

\clearpage
\newpage
\section{Placement Rules and Algorithms}
\label{sec:rules}
Table~\ref{table:placement_rules} outlines the placement rules employed by \methodname{} to govern the positioning of objects within a room. In addition to the parameter \texttt{x}, which specifies the buffer distance in meters, these rules also consider the dimensions of the object, the dimensions and positions of the anchor object (for rules like \texttt{place\_beside} that reference an anchor), and a mask of the room indicating its boundaries and spaces not occupied by previously placed objects. The LLM determines the appropriate rule for each object, along with these parameters. 

Algorithm.~\ref{alg:layout} details the furniture placement algorithm. The placement algorithm iteratively positions furniture groups $C_i$, starting with the anchor object for each group. This anchor is decided by the LLM. The anchor's placement utilizes its rule to identify viable spaces within the room (e.g., every available corner for \texttt{place\_corner}), placing the anchor where constraints are met. Should suitable spaces be unavailable, the furniture group is omitted. This approach is rational, as the LLM is instructed to prioritize the generation of the most significant furniture groups first (refer to Figure~\ref{fig:promtp2} for details), ensuring ample free space for their placement. Subsequently, the placement rule for each object within the group assesses all viable spaces that adhere to the constraints, such as positioning in front of the anchor, and discards any object for which no appropriate space is found. These rules not only guarantee that objects are placed in accordance with their relative positions within the room but also prevent objects from being positioned outside room boundaries or overlapping with one another.
\begin{algorithm}[h]
\caption{Furniture Placement Algorithm}
\label{alg:furniture-placement}
\begin{algorithmic}[1]
\renewcommand{\algorithmicrequire}{\textbf{Input:}}
\renewcommand{\algorithmicensure}{\textbf{Output:}}
\REQUIRE Room mask \( M_i \), 
Subgraphs \( C_i \), placement functions \( {P} \)
\STATE $B_i \gets \textit{empty}$
\FOR{subgraph \( c_i \) in \( C_i \)}
    \STATE /* Identify and place the anchor furniture */
    \STATE $n^{anchor}_{c_{i}} \leftarrow \text{identify\_anchor}(c_i)$
    \STATE $p \leftarrow  {P} ( n^{anchor}_{c_{i}})$
    \STATE $b^{anchor}_{c_{i}} \leftarrow p(M_i, \text{anchor})$
    \STATE $B_i \leftarrow \ B_i\ \cup \  b^{anchor}_{c_{i}}$
    \STATE \( M_i \leftarrow \text{update\_mask}(M_i, B_i) \)
    \STATE /* Place the subsequent furniture items */
    \FOR{each furniture item \( n^j_{c_{i}} \) in \( c_i \)}
        \STATE $p \leftarrow  {P}  ( n^j_{c_{i}})$
        \STATE $B_i \leftarrow \ B_i\ \cup \  p(M_i,  n^j_{c_{i}},b^{anchor}_{c_{i}})$
        \STATE \( M_i \leftarrow \text{update\_mask}(M_i, B_i) \)
    \ENDFOR
\ENDFOR
\ENSURE Set of furniture bounding boxes $B_i$
\end{algorithmic}
\label{alg:layout}
\end{algorithm}
\clearpage
\begin{table}[h]
    \begin{tabularx}{0.98\textwidth}{cX}
    \toprule
    \textbf{Rule} & \textbf{Description} \\
    \midrule
    \texttt{place\_corner} & Place the object in an available corner of the room. \\
    \addlinespace
    \texttt{place\_next\_wall} & Place the object next to an available wall section. \\
    \addlinespace
    \texttt{place\_wall} & Place the object against an available wall section, oriented towards the room center. \\
    \addlinespace
    \texttt{place\_center} & Place the object at the room center, or the center of the spare area that no objects have been placed above. \\
    \addlinespace
    \texttt{place\_next} & Place an anchor object next to an existing anchor object at the same orientation as that anchor. This is used for aligning groups of tables and chairs in rows and columns. \\
    \addlinespace
    \texttt{place\_beside(x)} & Place the object next to the anchor object with a buffer distance of x, such as a bed, at the same orientation as the anchor. This is used for object groups like a bed and two nightstands. \\
    \addlinespace
    \texttt{place\_around(x)} & Place the object at an available edge around the anchor object with a buffer distance of x, oriented towards the anchor. This is used for object groups like a dining table and chairs. \\
    \addlinespace
    \texttt{place\_front(x)} & Position the object in front of an anchor object with a buffer distance of x, oriented towards the anchor. This is used for object groups like a sofa and a TV stand. \\
    \addlinespace
    \texttt{place\_top(x)} & Place the object on top of an anchor object, with a buffer distance of x, at the center of the spare area on top of the anchor that no objects have been placed above. \\
    \addlinespace
    \texttt{place\_on\_wall(x)} & Place the object on an available wall section with a height of x, orientated towards the room center.\\
    
    \bottomrule
    \end{tabularx}
    \caption{The placement rules for room layout generation.}
    \label{table:placement_rules}
\end{table}
\clearpage
\newpage

\end{document}